\documentclass{article} % For LaTeX2e
\usepackage{iclr2019_conference,times}

\usepackage{amsmath,amsfonts,bm}

\def\eqref#1{equation~\ref{#1}}
\def\1{\bm{1}}

\def\vmu{{\bm{\mu}}}
\def\vtheta{{\bm{\theta}}}
\def\va{{\bm{a}}}

\def\vs{{\bm{s}}}

\def\vw{{\bm{w}}}
\def\vx{{\bm{x}}}

\DeclareMathAlphabet{\mathsfit}{\encodingdefault}{\sfdefault}{m}{sl}
\SetMathAlphabet{\mathsfit}{bold}{\encodingdefault}{\sfdefault}{bx}{n}

\newcommand{\E}{\mathbb{E}}

\newcommand{\R}{\mathbb{R}}

\newcommand{\KL}{D_{\mathrm{KL}}}

\usepackage{hyperref}
\usepackage{url}

\usepackage{graphicx}
\usepackage{subcaption}
\usepackage{algorithm}
\usepackage{algorithmicx}
\usepackage{algpseudocode}
\newcommand{\vect}[1]{{\boldsymbol{#1}}}

\newcommand{\intd}{\textrm{d} }

\newcommand{\target}{\textrm{target}}

\newcommand{\noise}{\textrm{noise}}
\newcommand{\Ad}{\textrm{Ad}}

\title{Hierarchical Reinforcement Learning \\ via Advantage-Weighted Information \\ Maximization}

\author{Takayuki Osa \\
University of Tokyo, Tokyo, Japan\\
RIKEN AIP, Tokyo, Japan\\
\texttt{osa@mfg.t.u-tokyo.ac.jp} \\
\And
Voot Tangkaratt\\
RIKEN AIP, Tokyo, Japan \\
\texttt{voot.tangkaratt@riken.jp} \\
\AND
Masashi Sugiyama \\
RIKEN AIP, Tokyo, Japan \\
University of Tokyo, Tokyo, Japan\\
\texttt{sugi@k.u-tokyo.ac.jp}
}

\iclrfinalcopy % Uncomment for camera-ready version, but NOT for submission.
\begin{document}

\maketitle

\begin{abstract}
Real-world tasks are often highly structured. Hierarchical reinforcement learning~(HRL) has attracted research interest as an approach for leveraging the hierarchical structure of a given task in reinforcement learning~(RL). 
However, identifying the hierarchical policy structure that enhances the performance of RL is not a trivial task. In this paper, we propose an HRL method that learns a latent variable of a hierarchical policy using mutual information maximization. 
Our approach can be interpreted as a way to learn a discrete and latent representation of the state-action space. 
To learn option policies that correspond to modes of the advantage function, we introduce \textit{advantage-weighted importance sampling}. 
In our HRL method, the gating policy learns to select option policies based on an option-value function, and these option policies are optimized based on the deterministic policy gradient method. This framework is derived by leveraging the analogy between a monolithic policy in standard RL and a hierarchical policy in HRL by using a deterministic option policy. 
Experimental results indicate that our HRL approach can learn a diversity of options and that it can enhance the performance of RL in continuous control tasks.
\end{abstract}

\section{Introduction}
Reinforcement learning~(RL) has been successfully applied to a variety of tasks, including board games~\citep{silver16}, robotic manipulation tasks~\citep{Levine16b}, and video games~\citep{Mnih15}.
Hierarchical reinforcement learning~(HRL) is a type of RL that leverages the hierarchical structure of a given task by learning a hierarchical policy~\citep{Sutton99,Dietterich00}. 
Past studies in this field have shown that HRL can solve challenging tasks in the video game domain~\citep{Vezhnevets17,Bacon17} and robotic manipulation~\citep{Daniel16,Osa18b}.
In HRL, lower-level policies, which are often referred to as \textit{option policies}, learn different behavior/control patterns, and 
the upper-level policy, which is often referred to as the \textit{gating policy}, learns to select option policies.
Recent studies have developed HRL methods using deep learning~\citep{Goodfellow16} and have shown that HRL can yield impressive performance for complex tasks~\citep{Bacon17,Frans18,Vezhnevets17,Haarnoja18}. 
However, identifying the hierarchical policy structure that yields efficient learning is not a trivial task, since
the problem involves learning a sufficient variety of types of behavior to solve a given task.
%In addition, recent HRL methods are often on-policy~\cite{Vezhnevets17,Bacon17}. 
%However, it is known that on-policy and off-policy methods show their superiority in different problem domains~\cite{Henderson18},
%and we believe that developing an off-policy HRL method is beneficial to solve complex tasks where off-policy methods are suitable.

In this study, we present an HRL method via the mutual information~(MI) maximization with advantage-weighted importance, which we refer to as adInfoHRL.
We formulate the problem of learning a latent variable in a hierarchical policy as one of learning discrete and interpretable representations of states and actions.
Ideally, each option policy should be located at separate modes of the advantage function.
To estimate the latent variable that corresponds to modes of the advantage function, we introduce advantage-weighted importance weights.
Our approach can be considered to divide the state-action space based on an information maximization criterion, and it learns option policies corresponding to each region of the state-action space.
We derive adInfoHRL as an HRL method based on deterministic option policies that are trained based on an extension of the deterministic policy gradient~\citep{Silver14,Fujimoto18}.
%Our experimental results show that our method outperforms state-of-the-art methods in tasks which have been often addressed by using non-hierarchical RL methods.
The contributions of this paper are twofold: 
\begin{enumerate}
	\item We propose the learning of a latent variable of a hierarchical policy as a discrete and hidden representation of the state-action space.
	To learn option policies that correspond to the modes of the advantage function, we introduce advantage-weighted importance.
	%One can consider that our approach divide the state-action space based on an information maximization criterion and conquer each of them.
	\item We propose an HRL method, where the option policies are optimized based on the deterministic policy gradient 
	and the gating policy selects the option that maximizes the expected return. 
	The experimental results show that our proposed method adInfoHRL can learn a diversity of options on continuous control tasks.
	Moreover, our approach can improve the performance of TD3 on such tasks as the Walker2d and Ant tasks in OpenAI Gym with MuJoco simulator.
%	\item We implemented our HRL approach based on Twin Delayed Deep Deterministic policy gradient algorithm~(TD3) proposed by \citet{Fujimoto18}. 
	
\end{enumerate}

\section{Background}
In this section, we formulate the problem of HRL in this paper and describe methods related to our proposal.

\subsection{Hierarchical Reinforcement Learning}
We consider tasks that can be modeled as a Markov decision process~(MDP), consisting of a state space $\mathcal{S}$, an action space $\mathcal{A}$, a reward function $r: \mathcal{S} \times \mathcal{A} \mapsto \R$,
an initial state distribution $\rho(\vs_0)$,
and a transition probability $p(\vs_{t+1}|\vs_t, \va_t)$ that defines the probability of transitioning from state $\vs_t$ and action $\va_t$ at time $t$ to next state $\vs_{t+1}$. 
The return is defined as $R_t = \sum^{T}_{i=t} \gamma^{i-t} r(\vs_i, \va_i) $, where $\gamma$ is a discount factor, and
policy $\pi(\va|\vs)$ is defined as the density of action $\va$ given state $\vs$.
%The state distribution induced by the policy $\pi$ is denoted by 
Let $d^{\pi}(\vs)= \sum^{T}_{t=0}\gamma^{t} p(\vs_t = \vs)$ denote the discounted visitation frequency induced by the policy $\pi$. 
The goal of reinforcement learning is to learn a policy that maximizes the expected return 
$J(\pi) = \E_{\vs_0, \va_0, ...}[R_0]$ where $\vs_0 \sim \rho(\vs_0), \va \sim \pi$ and $ \vs_{t+1} \sim p(\vs_{t+1}|\vs_t, \va_t)$.
%$\E_{\vs_0, \va_0, ...}[\cdot]$ indicates the expectation with respect to the samples induced by the policy $\pi$.
By defining the Q-function as $Q^{\pi}(\vs, \va) = \E_{\vs_0, \va_0, ...}[R_t | \vs_t = \vs, \va_t=\va]$,
the objective function of reinforcement learning can be rewritten as follows:
%We assume that an environment can be represented as a Markov decision process that is described by a set of states $\mathcal{S}$, a set of actions $\mathcal{A}$, 
%a reward function $r: \mathcal{A} \times \mathcal{S} \mapsto \Real$, a distribution of initial states $\rho(\vect{s})$,
%transition probability $p(\vect{s}_{t+1}|\vect{s}_t, \vect{a}_t)$, and a discount factor $\gamma \in [0, 1]$. 
%A policy$\pi(\vect{a}|\vect{s})$ is a density of action given a state.
%
%
%By following the notation in~\cite{silver16}, we denote the density at the state $\vect{s}'$ after transitioning for $t$ time steps from the state $\vect{s}$ 
%by $p(\vect{s}\rightarrow \vect{s}', t, \pi ) $.
%The (improper{\footnote{$d^{\pi}(\vect{s})$ is an improper distribution in the sense that $\int d^{\pi}(\vect{s}) \intd \vect{s}$ is not equal to 1. }}) discounted state distribution induced by a policy $\pi$ is then denoted by $d^{\pi}(\vect{s}) = \int_{\mathcal{S}} \sum_{\infty}^{i=t} \gamma^{i-t} \rho(\vect{s}) p(\vect{s}\rightarrow \vect{s}', t, \pi ) \intd \vect{s} $.
%We refer to a discounted sum of the future rewards as \textit{return}: $R_t = \sum^{\infty}_{i=t} \gamma^{i-t} r_i$.
%The goal of the agent is learn a policy that maximizes the expected return from the start state $J(\pi) = \E_{\vect{s}_1 \sim \rho(\vect{s})}[ R_1 | \vect{s}_1 ]$.
%Using the (improper) discounted state distribution $d^{\pi}(\vect{s})$, this objective can be rewritten  as
\begin{align}
J(\pi) %= \E_{(\vect{s}, \vect{a}) \sim d^{\pi}(\vect{s})\pi(\vect{a}|\vect{s})} \left[ r | \vect{s}_0 \right]
= \iint d^{\pi}(\vs) \pi( \va | \vs ) Q^{\pi}(\vs, \va)  \textrm{d}\va \textrm{d}\vs.
\label{eq:rl}
\end{align}
%The conditional expectation of the return $Q^{\pi}(\vect{s}, \vect{a}) = \E_{(\vect{s}, \vect{a}) \sim d^{\pi}(\vect{s})\pi(\vect{a}|\vect{s})}[R_t | \vect{s}_t, \vect{a}_t]$ is often referred to as the Q-function or the action-value function~\cite{Sutton98}. 

Herein, we consider hierarchical policy $\pi(\vect{a}|\vect{s}) = \sum_{o \in \mathcal{O}} \pi(o|\vect{s}) \pi(\vect{a}|\vect{s}, o)$,
%\begin{align}
%\pi(\vect{a}|\vect{s}) = \sum_{o \in \mathcal{O}} \pi(o|\vect{s}) \pi(\vect{a}|\vect{s}, o),
%\end{align}
where $o$ is the latent variable and $\mathcal{O}$ is the set of possible values of $o$.
Many existing HRL methods employ a policy structure of this form~\citep{Frans18,Vezhnevets17,Bacon17,Florensa17,Daniel16}.
In general, latent variable $o$ can be discrete~\citep{Frans18,Bacon17,Florensa17,Daniel16,Osa18} or continuous~\citep{Vezhnevets17}.
$ \pi(o|\vect{s})$ is often referred to as a \textit{gating policy}~\citep{Daniel16, Osa18}, \textit{policy over options}~\citep{Bacon17}, or \textit{manager}~\citep{Vezhnevets17}. 
Likewise, $\pi(\vect{a}|\vect{s}, o)$ is often referred to as an \textit{option policy}~\citep{Osa18}, \textit{sub-policy}~\citep{Daniel16}, or \textit{worker}~\citep{Vezhnevets17}.
In HRL, the objective function is given by
\begin{align}
J(\pi) 
=  \iint d^{\pi}(\vect{s}) \sum_{o  \in \mathcal{O}} \pi( o | \vect{s} ) \pi( \vect{a} | \vect{s}, o ) Q^{\pi}( \vect{s}, \vect{a} ) \textrm{d}\vect{a}\textrm{d}\vect{s} .
\end{align}
As discussed in the literature on inverse RL~\citep{Ziebart10}, multiple policies can yield equivalent expected returns.
This indicates that there exist multiple solutions to latent variable $o$ that maximizes the expected return.
To obtain the preferable solution for $o$, we need to impose additional constraints in HRL.
Although prior work has employed regularizers~\citep{Bacon17} and constraints~\citep{Daniel16} to obtain various option policies, the method of learning  a \textit{good} latent variable $o$ that improves sample-efficiency of the learning process remains unclear. 
In this study we propose the learning of the latent variable by maximizing MI between latent variables and state-action pairs. 

\subsection{Deterministic Policy Gradient}
The deterministic policy gradient~(DPG) algorithm was developed for learning a monolithic deterministic policy 
$\vect{\mu}_{\vect{\theta}}(\vect{s}): \mathcal{S} \mapsto \mathcal{A}$ by \citet{Silver14}.
In off-policy RL, the objective is to maximize the expectation of the return, averaged over the state distribution induced by a behavior policy $\beta(\vect{a}|\vect{s})$:
\begin{align}
J(\pi) % = \int d^{\beta}(\vect{s}) V^{\pi}\big(\vect{s}) \intd\vect{s}  
= \iint d^{\beta}(\vect{s}) \pi( \vect{a} | \vect{s}) Q^{\pi}\big(\vect{s}, \vect{a} ) \intd\vect{a}\intd\vect{s}. 
\label{eq:offRL}
\end{align}
When a policy is deterministic, the objective becomes 
$J(\pi) =  \int d^{\beta}(\vect{s}) Q^{\pi}\big(\vect{s}, \vect{\mu}_{\vect{\theta}}(\vect{s} ) \big) \intd\vect{s}$.
%\begin{align}
%J(\pi) =  \int d^{\beta}(\vect{s}) Q^{\pi}\big(\vect{s}, \vect{\mu}_{\vect{\theta}}(\vect{s} ) \big) \intd\vect{s}.
%\label{eq:offRL_determ}
%\end{align}
\citet{Silver14} have shown that the gradient of a deterministic policy is given by
\begin{align}
\nabla_{\vtheta} \E_{\vs \sim d^{\beta}(\vs)}[ Q^{\pi}(\vs, \va)]  
= \E_{\vs \sim d^{\beta}(\vs)} \left[ \nabla_{\vtheta} \vmu_{\vtheta}(\vs) \nabla_{\va} Q^{\pi}\big(\vs, \va \big) |_{\va = \vmu_{\vtheta}(\vs)} \right].
\end{align}
The DPG algorithm has been extended to the deep deterministic policy gradient~(DDPG) for continuous control problems that require neural network policies~\citep{Lillicrap15}.
Twin Delayed Deep Deterministic policy gradient algorithm~(TD3) proposed by \citet{Fujimoto18} is a variant of DDPG that outperforms the state-of-the-art on-policy methods such as TRPO~\citep{Schulman17} and PPO~\citep{Schulman2017_ppo} in certain domains.
We extend this deterministic policy gradient to learn a hierarchical policy.

\subsection{Representation Learning via Information Maximization}
\label{sec:rim}
Recent studies such as those by \citet{Xi16,Hu17,Li17} have shown that an interpretable representation can be learned by maximizing MI.
Given a dataset $X=(\vect{x}_1, ..., \vect{x}_n)$, 
regularized information maximization~(RIM) proposed by \citet{Gomes10} involves learning a conditional model $\hat{p}(y| \vx; \vect{\eta})$ with parameter vector $\vect{\eta}$ 
that predicts a label $y$.
The objective of RIM is to minimize
\begin{align}
\ell(\vect{\eta}) - \lambda I_{\vect{\eta}}(\vx, y),
\end{align}
where $\ell(\vect{\eta})$ is the regularization term, $I_{\vect{\eta}}(\vx, y)$ is MI, and $\lambda$ is a coefficient. MI can be decomposed as 
$I_{\vect{\eta}}(\vx, y) = H(y)  - H(y|\vx)$ where $H(y)$ is entropy and $H(y|\vx)$ the conditional entropy.
%Prior work~\cite{Gomes10,Hu17} employed the empirical estimate of the mutual information given by $\hat{I}_{\vect{w}}(y; \vx) = \hat{H}(y)  - \hat{H}(y|\vx)$
%where $\hat{H}(y) = - \hat{p}(y; \vw) \log \hat{p}(y; \vw)$ and $\hat{p}(y; \vw)$ is 
%the empirical estimate of $p(y)= \int p(\vx) p(y | \vx) \intd \vx = \E[ p(y | \vx) ]$ given by
%\begin{align}
%\hat{p}(y; \vw) = \frac{1}{N} \sum_{i=1}^{N} \hat{p}(y_i | \vx_i; \vw).
%\end{align}
%Likewise, $\hat{H}(y|\vx)$ is the empirical estimates of $H(y|\vx)= - \E \left[ p( y | \vx) \log p( y | \vx) \right]$ 
%given by 
%\begin{align}
%\hat{H}(y|\vx) = - \frac{1}{N} \sum_{i}^{N} \hat{p}(y_i | \vx_i; \vect{w}) \log \hat{p}( y_i | \vx_i; \vw).
%\end{align}
Increasing $H(y)$ conduces the label to be uniformly distributed, and decreasing $H(y|\vx)$ conduces to clear cluster assignments.
Although RIM was originally developed for unsupervised clustering problems,
the concept is applicable to various problems that require learning a hidden discrete representation.
In this study, we formulate the problem of learning the latent variable $o$ of a hierarchical policy as one of learning a latent representation of the state-action space.

\section{Learning Options via Advantage-Weighted Information Maximization}
In this section, we propose a novel HRL method based on advantage-weighted information maximization.
We first introduce the latent representation learning via advantage-weighted information maximization, 
and we then describe the HRL framework based on deterministic option policies.

%\paragraph{Finding the modes of Advantage function}
\subsection{Latent Representation Learning via Advantage-Weighted Information Maximization} 
\begin{figure}[]
	\centering
	\begin{subfigure}[t]{0.27\columnwidth}
		\includegraphics[width=\textwidth]{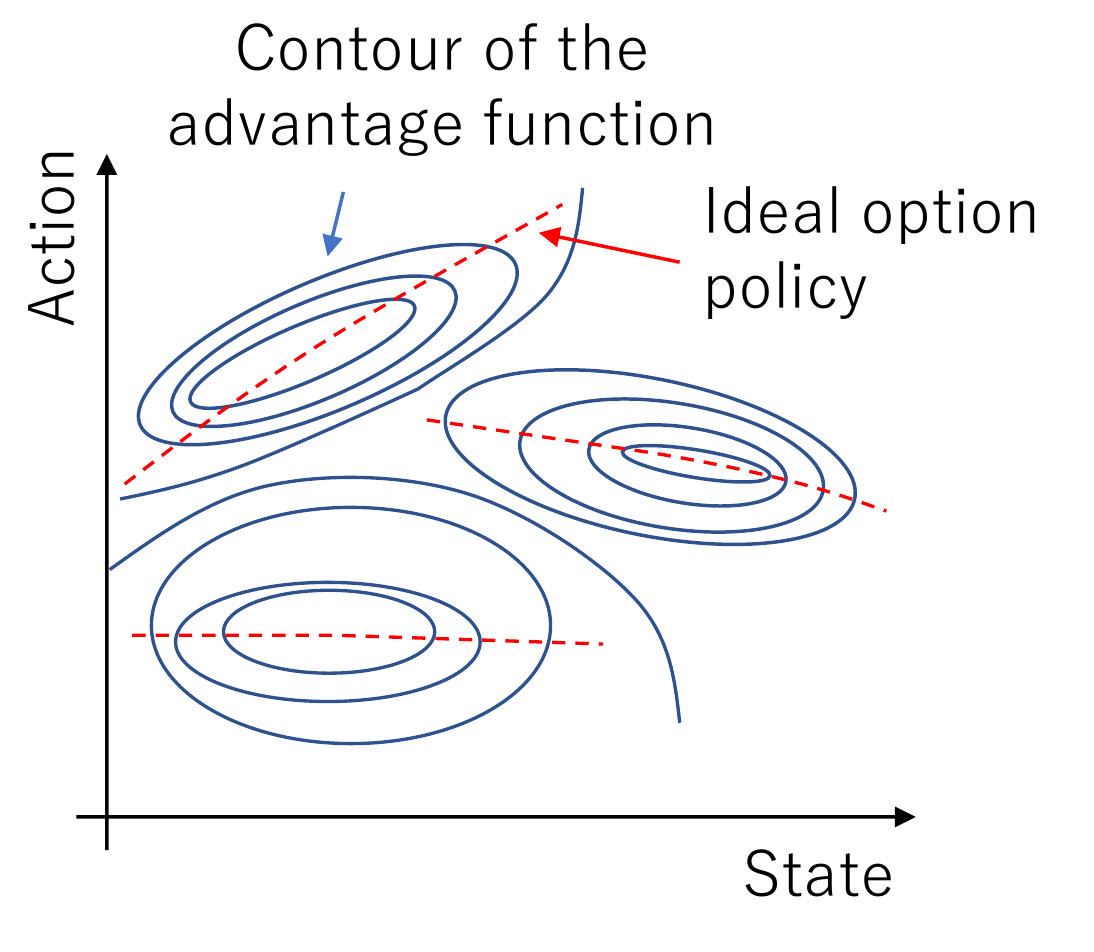}
		\caption{Visualization of the advantage function with multiple modes in the state-action space.}
	\end{subfigure}
	\begin{subfigure}[t]{0.27\columnwidth}
		\includegraphics[width=\textwidth]{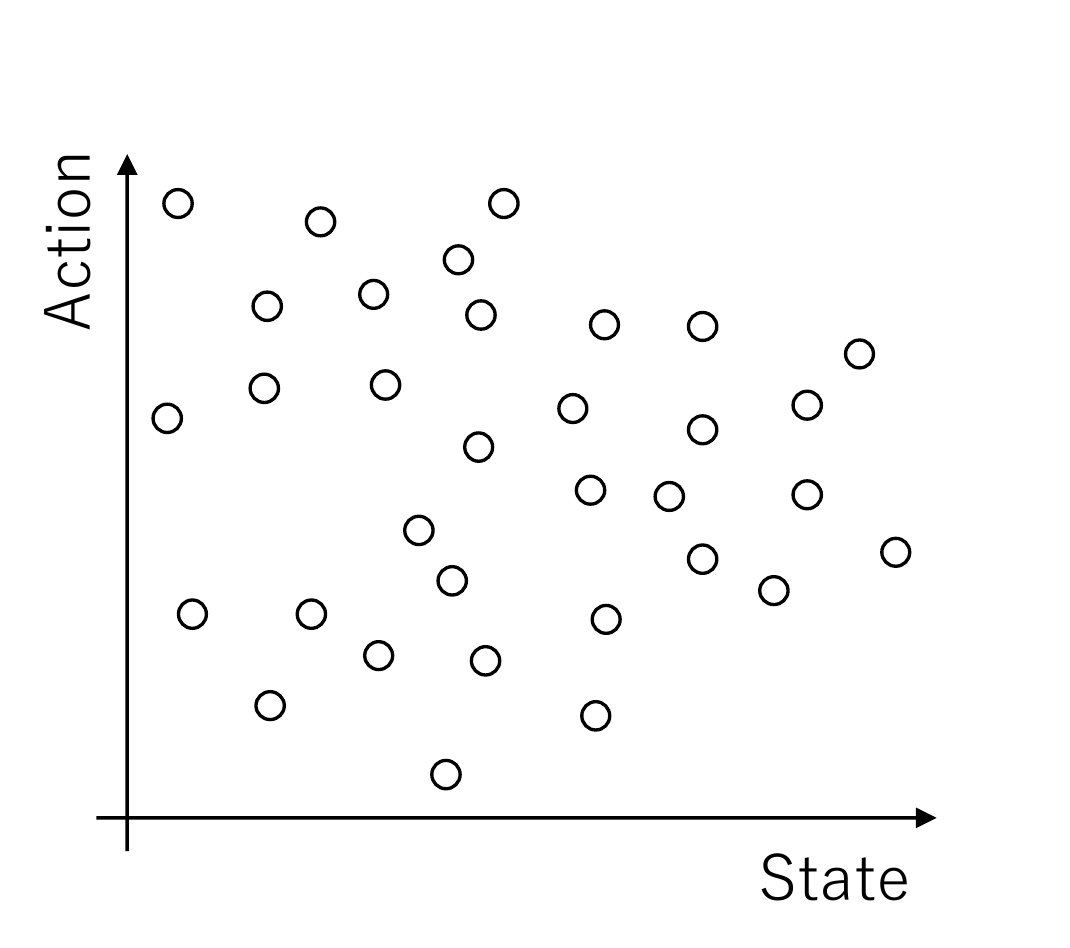}
		\caption{Modes of the density induced by arbitrary policies do not correspond to the modes of the advantage function.}
	\end{subfigure}
	\begin{subfigure}[t]{0.27\columnwidth}
		\includegraphics[width=\textwidth]{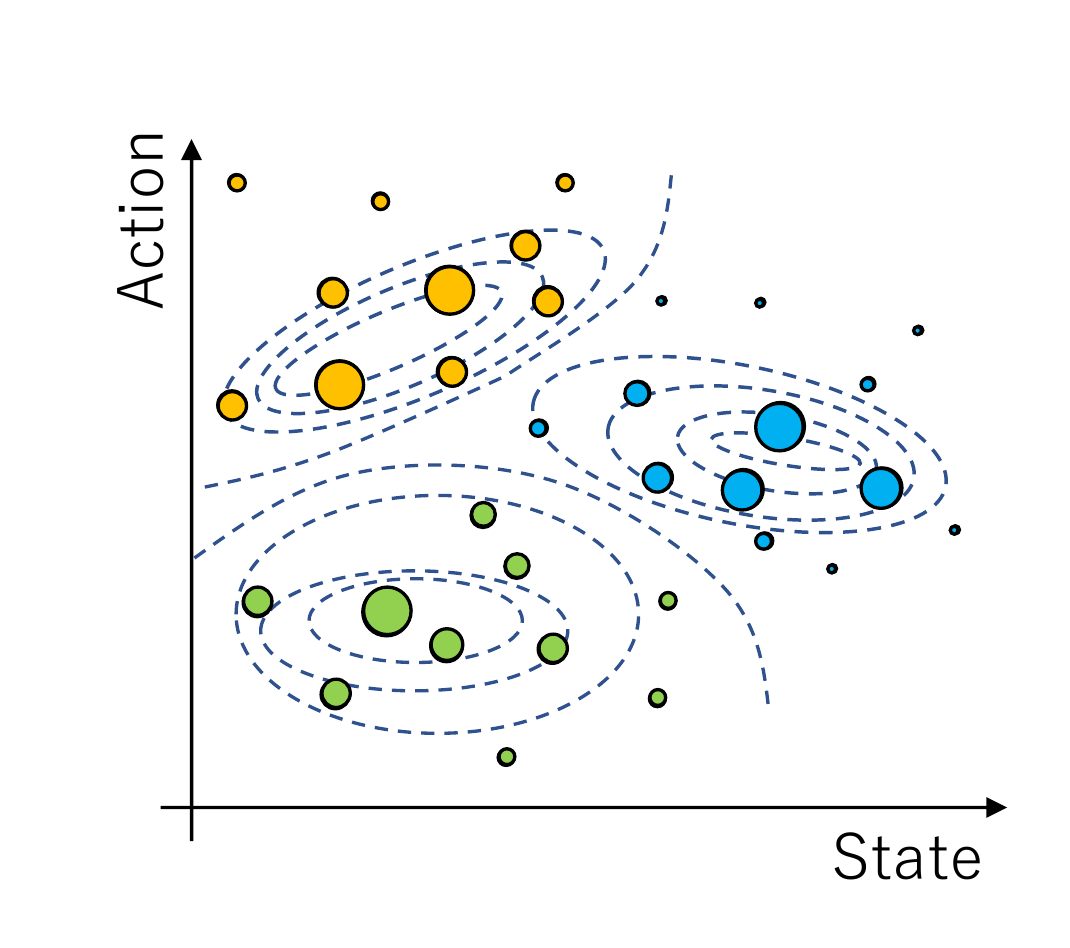}
		\caption{Modes of the density estimated with advantage-weighted importance correspond to the modes of the advantage function.}
	\end{subfigure}
	\caption{Schematic sketch of our HRL approach. 
		By using the advantage-weighted importance, 
		the problem of finding the modes of the advantage-function can be reduced to that of finding the modes of the density of state action pairs.
	}
	\label{fig:schematic}
\end{figure}

Although prior work has often considered $H(o|\vs)$ or $I(\vs, o)$, which results in a division of the state space, 
we are interested in using $I\big((\vs, \va), o\big)$ for dividing the state-action space instead.
A schematic sketch of our approach is shown in Figure~\ref{fig:schematic}.
As shown in the left side of Figure~\ref{fig:schematic}, the advantage function often has multiple modes. 
Ideally, each option policies should correspond to separate modes of the advantage function.
However, it is non-trivial to find the modes of the advantage function in practice. 
For this purpose, we reduce the problem of finding modes of the advantage function to that of finding the modes of the probability density of state action pairs.

We consider a  policy based on the advantage function of the form
\begin{align}
\pi_{\Ad}(\va | \vs) = \frac{ f \big(A^{\pi}(\vect{s}, \vect{a}) \big)}{Z},
\end{align}
where $A^{\pi}(\vect{s}, \vect{a})=Q^{\pi}(\vect{s}, \vect{a}) - V^{\pi}(\vs)$ is the advantage function, $V^{\pi}(\vs)$ is the state value function, and $Z$ is the partition function.  
$f(\cdot)$ is a \textit{functional}, which is a function of a function.
$f(\cdot)$ is a monotonically increasing function with respect to the input variable and always satisfies $f(\cdot) > 0$. 
In our implementation we used the exponential function $f(\cdot) = \exp(\cdot)$.
%A policy with this form has been often employed in the RL literature~\citep{Akrour16, Deisenroth13}.
When following such a policy, an action with the larger advantage is drawn with a higher probability.
Under this assumption, finding the modes of the advantage function is equivalent to finding modes of the density induced by $\pi_{\Ad}$.
Thus, finding the modes of the advantage function can be reduced to the problem of clustering samples induced by $\pi_{\Ad}$.

Following the formulation of RIM introduced in Section~\ref{sec:rim}, we formulate the problem of clustering samples induced by $\pi_{\Ad}$ as the learning of discrete representations via MI maximization.
For this purpose, we consider a neural network that estimates $p(o|\vect{s},\vect{a}; \vect{\eta})$ parameterized with vector $\vect{\eta}$, which we refer to as the \textit{option network}.
We formulate the learning of the latent variable $o$ as minimizing 
\begin{align}
\mathcal{L}_{\textrm{option}}(\vect{\eta}) =  \ell(\vect{\eta}) - \lambda I\big( o, (\vs, \va); \vect{\eta} \big) ,
\label{eq:option_loss}
\end{align}
where $I( o, (\vs, \va) ) = \hat{H}(o| \vs, \va ; \vect{\eta}) -  \hat{H}(o; \vect{\eta})$,
and $\ell(\vect{\eta})$ is the regularization term. 
In practice, we need to approximate the advantage function, and we learn the discrete variable $o$ that corresponds to the modes of the current estimate of the advantage function.
For regularization, we used a simplified version of virtual adversarial training~(VAT) proposed by \citet{Miyato16}.
Namely, we set $\ell(\vect{\eta}) = \KL\big( p(o| \vs^{\noise}, \va^{\noise}; \vect{\eta}) || p(o|\vs, \va; \vect{\eta}) \big)$ where $\vs^{\noise} = \vs + \vect{\epsilon}_\vs$, $\va^{\noise} = \va + \vect{\epsilon}_\va$, $\vect{\epsilon}_\vs$ and $\vect{\epsilon}_\va$ denote white noise.
This regularization term penalizes dissimilarity between an original state-action pair and a perturbed one,
and \citet{Hu17} empirically show that this regularization improves the performance of learning latent discrete representations.

When computing MI, we need to compute $p(o)$ and $H(o|\vs, \va)$ given by
\begin{align}
p(o) & = \int p^{\pi_{\Ad}}(\vect{s}, \vect{a})  p(o | \vect{s}, \vect{a}; \vect{\eta}) \intd\vect{a}\intd\vect{s} 
= \E_{(\vect{s}, \vect{a}) \sim p^{\pi_{\Ad}}(\vect{s}, \vect{a})} \left[ p(o | \vect{s}, \vect{a}; \vect{\eta})  \right]\\
%\end{align}
%and
%\begin{align}
H(o|\vs, \va) %& = \int p^{\pi_{\Ad}}(\vs, \va) p(o | \vs, \va; \vect{\eta}) \log p(o|\vs, \va; \vect{\eta}) \intd\vs\intd\va \\
& = \E_{(\vect{s}, \vect{a}) \sim p^{\pi_{\Ad}}(\vs, \va)} \left[ p(o | \vs, \va; \vect{\eta}) \log p(o | \vs, \va; \vect{\eta}) \right].
\end{align}
Thus, the probability density of $(\vs, \va)$ induced by $\pi_{\Ad}$ is necessary for computing MI for our purpose. To estimate the probability density of $(\vs, \va)$ induced by $\pi_{\Ad}$, we introduce the advantage-weighted importance in the next section.

\subsection{Importance Weights for Mutual Information Estimation}
Although we show that the problem of finding the modes of the advantage function can be reduced to MI maximization with respect to the samples induced by $\pi_{\Ad}$, samples induced by $\pi_{\Ad}$ are not available in practice.
While those induced during the learning process are available, a discrete representation obtained from such samples does not correspond to the modes of the advantage function.  
To estimate the density induced by $\pi_{\Ad}$, we employ an importance sampling approach.
%While using soft actor critic (SAC)~\citep{Haarnoja18} can also be used for generate the samples that follows $\pi_{\textrm{Ad}}$,
%we employ an importance sampling approach in this study because it can be used for HRL methods which are not based on SAC.
%add some reference or give some reasons of using importance sampling

We assume that the change of the state distribution induced by the policy update is sufficiently small, namely, $d^{\pi_{\Ad}}(\vect{s}) \approx d^{\beta}(\vect{s})$. 
Then, the importance weight can be approximated as 
\begin{align}
W( \vect{s}, \vect{a})  = \frac{p^{\pi_{\Ad}}( \vect{s}, \vect{a})}{p^{\beta}( \vs, \va)}  = \frac{ d^{\pi_{\Ad}}(\vect{s}) \pi_{\Ad}(\vect{a}|\vect{s}) }{d^{\beta}(\vect{s})\beta(\vect{a} | \vect{s} )} 
\approx \frac{\pi_{\Ad}(\vect{a}|\vect{s})}{\beta(\vect{a} | \vect{s} )} 
= \frac{f(A(\vect{s}, \vect{a}) )}{Z \beta(\va | \vs)}.
\end{align}
%When normalizing the importance weight, the partition function $Z$ is canceled.
and the normalized importance weight is given gy
\begin{align}
\tilde{W}( \vs, \va)  = \frac{W( \vs, \va)}{ \sum_{j=1}^{N} W( \vs_j, \va_j) }  
= \frac{ \frac{f(A(\vs, \va) )}{Z \beta(\va | \vs)} }{ \sum_{j=1}^{N} \frac{f(A(\vs_j, \va_j) )}{Z \beta(\va_j | \vs_j)}} 
= \frac{ \frac{f(A(\vs, \va) )}{\beta(\va | \vs)} }{ \sum_{j=1}^{N} \frac{f(A(\vs_j, \va_j) )}{ \beta(\va_j | \vs_j)}}.
\end{align}
As the partition function $Z$ is canceled, we do not need to compute $Z$ when computing the importance weight in practice.
We call this importance weight $W$ the \textit{advantage-weighted importance} and employ it to compute the objective function used to estimate the latent variable.

This advantage-weighted importance is used to compute the entropy terms for computing MI in Equation~(\ref{eq:option_loss}).
The empirical estimate of the entropy $ H(o)$ is given by 
\begin{align}
\hat{H}(o; \vect{\eta}) = - \sum_{o \in \mathcal{O}} \hat{p}(o; \vect{\eta}) \log \hat{p}(o; \vect{\eta}), \textrm{where} \
\hat{p}(o; \vect{\eta}) = \frac{1}{N} \sum_{i=1}^{N} W(\vs_i, \va_i) p(o | \vs_i, \va_i; \vect{\eta}).
\label{eq:weighted_p}
\end{align}
where the samples $(\vs_i, \va_i)$ are drawn from $p^{\beta}(\vs, \va)$ induced by a behavior policy $\beta(\va|\vs)$.
Likewise, the empirical estimate of the conditional entropy $H(o|\vs, \va) $ is given by
\begin{align}
\hat{H}(o|\vs, \va; \vect{\eta}) 
%& = \int p^{\beta}(\vect{s}, \vect{a}) p(o | \vect{s}, \vect{a}) \log p(o|\vect{s}, \vect{a}) \intd\vect{s}\intd\vect{a} \\
%& = \E \left[  p(o | \vect{s}, \vect{a}) \log p(o | \vect{s}, \vect{a}) \right] 
= \frac{1}{N} \sum_{i}^{N}  W(\vs_i, \va_i) p(o | \vect{s}_i, \vect{a}_i; \vect{\eta}) \log p(o | \vect{s}_i, \vect{a}_i; \vect{\eta}).
\label{eq:weighted_entropy}
\end{align}
The derivations of Equations (\ref{eq:weighted_p}) and (\ref{eq:weighted_entropy}) are provided in Appendix~\ref{app:weighted_info}.
To train the option network, we store the samples collected by the $M$ most recent behavior policies, to which we refer as on-policy buffer $D_{\textrm{on}}$.
Although the algorithm works with entire samples stored in the replay buffer, 
we observe that the use of the on-policy buffer for latent representation learning exhibits better performance. 
For this reason, we decided to use the on-policy buffer in our implementation.
Therefore, while the algorithm is off-policy in the sense that the option is learned from samples collected by behavior policies,
our implementation is ``semi''on-policy in the sense that we use samples collected by the most recent behavior policies.

\section{HRL Objective with Deterministic Option Policies}
Instead of stochastic option policies, we consider deterministic option policies and model them using separate neural networks. 
We denote by $\pi(\vect{a}|\vect{s}, o) = \vmu^o_{\vtheta}(\vs)$ deterministic option policies parameterized by vector $\vtheta$.
The objective function of off-policy HRL with deterministic option policies can then be obtained by replacing $\pi(\va|\vs)$ with $\sum_{o  \in \mathcal{O}}  \pi( o | \vs ) \pi( \va | \vs, o )$ in Equation~(\ref{eq:offRL}):
\begin{align}
J(\vw, \vtheta) 
%& = \iint d^{\beta}(\vs) \sum_{o \in \mathcal{O}} d^{\pi}(\vs)  \pi( o | \vs ) \pi( \va | \vs, o ) Q^{\pi}\big( \vs, \va ) \big) \intd\vect{a}\intd\vect{s} \\
= \int d^{\beta}(\vs) \sum_{o  \in \mathcal{O}} \pi( o | \vs ) Q^{\pi}\big(\vs, \vmu^o_{\vtheta}(\vs); \vw \big) \intd\vs,
\label{eq:offHRL}
\end{align}
where $Q^{\pi}(\vs, \va; \vw)$ is an approximated Q-function parameterized using vector $\vw$.
This form of the objective function is analogous to Equation~(\ref{eq:offRL}). 
Thus, we can extend standard RL techniques to the learning of the gating policy $\pi(o | \vect{s})$ in HRL with deterministic option policies.

In HRL, the goal of the gating  policy is to generate a value of $o$ that maximizes the conditional expectation of the return:
\begin{align}
Q^{\pi}_{\Omega}(\vect{s}, o) = \E \left[ R | \vect{s}_t = \vs, o_t=o \right] = \int \pi(\vect{a} | \vect{s}, o ) Q^{\pi}(\vect{s}, \vect{a}) \intd \vect{a},
\end{align}
which is often referred to as the option-value function~\citep{Sutton99}.
When option policies are stochastic, it is often necessary to approximate the option-value function $Q^{\pi}_{\Omega}(\vs, o)$ in addition to the action-value function $Q^{\pi}(\vs, \va)$.
However, in our case, the option-value function for deterministic option policies is given by
\begin{align}
Q^{\pi}_{\Omega}(\vs, o) =  Q^{\pi}(\vect{s}, \vmu^o_{\vtheta}(\vs)),
\end{align}
which we can estimate using the deterministic option policy $\vmu^o_{\vtheta}(\vs)$ and the approximated action-value function $Q^{\pi}(\vs, \va; \vw)$.
%\paragraph{Gating Policy}
%In this work, we formulate the gating policy by following the maximum entropy~~(MaxEnt) RL~\cite{Ziebart10,Haarnoja17,Schulman17}.
%The MaxEnt RL encodes stochasticity in its policy form and exhibits good exploration performance~\cite{Haarnoja17}.
%In our method, the MaxEnt RL framework provides efficient exploration across different options.
In this work we employ the softmax gating policy of the form
\begin{align}
\pi(o | \vs) % = \softmax\limits_{o} Q^{\pi}(\vect{s}, \vect{\mu}_{\vtheta}}(\vect{s}, o)) 
= \frac{\exp\big(Q^{\pi}(\vs, \vmu^{o}_{\vtheta}(\vs); \vw ) \big)}{\sum_{o \in \mathcal{O}} \exp\left(Q^{\pi}\big(\vs, \vmu^{o}_{\vtheta}(\vs) ; \vw \big) \right) },
\label{eq:gate}
\end{align}
which encodes the exploration in its form~\citep{Daniel16}.
The state value function is given as
\begin{align}
V^{\pi}(\vs) = \sum_{o \in \mathcal{O}} \pi(o|\vs)  Q^{\pi}(\vect{s}, \vmu^{o}_{\vtheta}(\vs); \vw ),
\end{align}
which can be computed using Equation (\ref{eq:gate}). We use this state-value function when computing the advantage-weighted importance as $A(\vs, \va) = Q(\vs, \va) - V(\vs)$.
%\begin{align}
%V^{\pi}(\vect{s}) = \E_{o\sim q(o)} \left[ \frac{\pi(o|\vect{s}) \big( Q^{\pi}(\vect{s}, \vect{\mu}^{o}_{\vect{\theta}}(\vect{s}); \vect{w} ) \big)}{q(o)} \right],
%\end{align}
%where $q$ is an arbitrary probability mass function for sampling $o$.
In this study, the Q-function is trained in a manner proposed by \citet{Fujimoto18}.
Two neural networks $(Q^{\pi}_{\vw_1}, Q^{\pi}_{\vw_2})$ are trained to estimate the Q-function, 
and the target value of the Q-function is computed as $y_i = r_i + \gamma \min_{1, 2} Q(\vs_i, \va_i)$ for sample $(\vs_i, \va_i, \va'_i, r_i)$ in a batch sampled from a replay buffer, where $r_i = r(\vs_i, \va_i)$.
%which is analogous to the soft Q-learning~\cite{Haarnoja17,Schulman17} for a monolithic policy.
%The target value of the action-value function is given by 
%$\bar{Q}^{\pi_{\vect{\theta}}}(\vect{s}, \vect{a}) = r + \gamma  V^{\pi}_{\textrm{soft}}(\vect{s}')$,
%where $V^{\pi}_{\textrm{soft}}(\vect{s}) $ is the soft state value function given by
%\begin{align}
%V^{\pi}_{\textrm{soft}}(\vect{s}) = \log \E_{o\sim q(o)} \left[ \frac{ \frac{1}{\alpha} \exp \big( Q^{\pi}(\vect{s}, \vect{\mu}_{\vect{\theta}}(\vect{s}, o); \vect{w} ) \big)}{q(o)} \right],
%\end{align}
%and $\alpha$ is the partition function and $q$ is an arbitrary probability mass function for sampling $o$.
In this study, the gating policy determines the option once every $N$ time steps, i.e., $t = 0, N, 2N,\ldots$
%Given the data samples induced by the behavior policy $\beta$, namely $\{ (\vect{s}_i, \vect{a}_i, r_i) \}_{i}^{N} \sim d^{\beta}(\vect{s})\beta(\vect{a}|\vect{s})$, 

%\begin{figure}[]
%	\centering
%	\begin{subfigure}[b]{0.33\columnwidth}
%		\includegraphics[width=\textwidth]{critic}
%		\caption{Q network.}
%		%\label{fig:net_structure1}
%	\end{subfigure}
%	\begin{subfigure}[b]{0.32\columnwidth}
%		\includegraphics[width=\textwidth]{option}
%		\caption{Option network.}
%		%\label{fig:net_structure2}
%	\end{subfigure}
%	\begin{subfigure}[b]{0.28\columnwidth}
%		\includegraphics[width=\textwidth]{actor}
%		\caption{Option-policy network.}
%		%\label{fig:net_structure2}
%	\end{subfigure}
%	\caption{Network architectures in adInfoHRL. 
%		In a rollout, given a state $\vect{s}$, $o$ is drawn by following the gating policy in~\eqref{eq:gate} using both the option-Q and option-policy networks. 
%		Actions are generated with the option-policy network using $\vect{s}$ and $o$ drawn by the gating policy.}
%	\label{fig:net_structure}
%\end{figure}

%\paragraph{Deterministic Policy Gradient for Option Policies} 
Neural networks that model $\vmu^{o}_{\vtheta}(\va|\vs)$ for $o=1,...,O$, which we refer to as \textit{option-policy networks}, are trained separately for each option.
In the learning phase, $p(o|\vs, \va)$ is estimated by the option network. 
Then, samples are assigned to option $o^{*} = \arg \max_o p(o|\vs, \va; \vect{\eta})$ and are used to update the option-policy network that corresponds to $o^{*}$.
When performing a rollout, $o$ is drawn by following the gating policy in Equation~(\ref{eq:gate}), and an action is generated by the selected option-policy network.
%The network structure is illustrated in Figure~\ref{fig:net_structure}.

%We learn option policies $\vmu^{o}_{\vtheta}(\vs)$ for $o=1,\ldots,O$ to maximize the conditional expectation of the return given $o$, 
%with respect to the state distribution induced by the behavior policy $\beta(\vect{a}|\vect{s})$:
%\begin{align}
%\E_{\vs \sim d^{\beta}(\vs)}[ Q^{\pi}(\vs, \va) | o_t = o ] = \int d^{\beta}(\vs) Q^{\pi}\big(\vs, \vmu^{o}_{\vtheta}(\vs) \big) \intd\vect{s}.
%\end{align}
%It is worth noting that $\E_{d^{\beta}}[ Q^{\pi}(\vs, \va) | o_t = o ]$ is the sum of expected rewards obtained by using the option policy $\vmu^{o}_{\vtheta}(\vs)$ at time $t$ and following the gating policy $\pi(o|\vs)$ and option policies $\vmu^{o}_{\vtheta}(\vs)$, which is \textit{not} the sum of expected rewards obtained by using the option policy $\vmu^{o}_\vtheta(\vs)$ until the end of the time horizon.

Differentiating the objective function in Equation~(\ref{eq:offHRL}), we obtain the deterministic policy gradient of our option-policy $\vmu^{o}_{\vtheta}(\vs)$ given by  
\begin{align}
\nabla_{\vect{\theta}} J(\vect{w}, \vect{\theta}) = \E_{\vect{s} \sim d^{\beta}(\vect{s})\pi(o|\vs)} \left[ \nabla_{\vect{\theta}} \vect{\mu}^{o}_{\vect{\theta}}(\vect{s}) \nabla_{\vect{a}} Q^{\pi}\big(\vect{s}, \vect{a} \big) |_{\vect{a} = \vect{\mu}^{o}_{\vect{\theta}}(\vect{s})} \right].
\label{eq:dpg_option}
\end{align}
%\begin{align}
%&\nabla_{\vect{\theta}} \E_{\vect{s} \sim d^{\beta}(\vect{s})}[ Q^{\pi}(\vect{s}, \vect{a}) | o_t = o ] = \E_{\vect{s} \sim d^{\beta}(\vect{s})} \left[ \nabla_{\vect{\theta}} \vect{\mu}^{o}_{\vect{\theta}}(\vect{s}) \nabla_{\vect{a}} Q^{\pi}\big(\vect{s}, \vect{a} \big) |_{\vect{a} = \vect{\mu}^{o}_{\vect{\theta}}(\vect{s})} \right].
%\end{align}

The procedure of adInfoHRL is summarized by Algorithm~\ref{alg:adInfoHRL}.
As in TD3~\citep{Fujimoto18}, we employed the soft update using a target value network and a target policy network.

\begin{algorithm}[t]
	\caption{ HRL via Advantage-Weighted Information Maximization~(adInfoHRL) }
	\begin{algorithmic}
		\State{\textbf{Input:} Number of options $O$, size of on-policy buffer}
		\State{\textbf{Initialize:} Replay buffer $\mathcal{D}_{R}$, on-policy buffer $\mathcal{D}_{\textrm{on}}$, network parameters $\vect{\eta}$, $\vect{\theta}$, $\vect{w}$, $\vect{\theta}^{\target}$, $\vect{w}^{\target}$  }
		\Repeat{}
		\For{$t=0$ to $t=T$}
		\State{Draw an option for a given $\vs$ by following Equation~\ref{eq:gate}: $o \sim \pi(o|\vs)$}
%		\State{$o \sim \pi(o|\vect{s}) = \textrm{softmax}_{o} Q(\vect{s}, \vect{\mu}(\vect{s}, o))$}
		\State{Draw an action $\va \sim \beta(\va|\vs, o) = \vmu^{o}_{\vtheta}(\vect{s}) + \vect{\epsilon}$}
		\State{Record a data sample $(\vs, \va,  r, \vs') $ }
		\State{Aggregate the data in $\mathcal{D}_{R}$ and $\mathcal{D}_{\textrm{on}}$}
		%\State{Collect samples $\mathcal{D}_{\textrm{new}} = \{ (\vect{a}^{\new}_i, \vect{s}^{\new}_i, r^{\new}_i) \}^{N}_{i=1} $}
		\If{the on-policy buffer is full}
		\State{Update the option network by minimizing Equation~(\ref{eq:option_loss}) for samples in $\mathcal{D}_{\textrm{on}}$ }
		\State{Clear the on-policy buffer $\mathcal{D}_{\textrm{on}}$}
		\EndIf
		\State{Sample a batch $\mathcal{D}_{\textrm{batch}} \in \mathcal{D}_{R} $}
		\State{Update the Q network parameter $\vw$}
		\If{t mod d}
		\State{Estimate $p(o | \vect{s}_i, \vect{a}_i) $ for $( \vect{s}_i, \vect{a}_i) \in \mathcal{D}_{\textrm{batch}}$ using the option network}
		\State{Assign samples $( \vect{s}_i, \vect{a}_i) \in \mathcal{D}_{\textrm{batch}}$ to the option $o^{*} = \arg \max p(o| \vs_i, \va_i)$}
		\State{Update the option policy networks $\vect{\mu}^{o}_{\vect{\theta}}(\vect{s})$ for $o=1,...,O$ with Equation~(\ref{eq:dpg_option}) }
		%using $\{ (\vect{s}_i, \vect{a}_i, , r_i, \vect{s}'_i, p(o | \vect{s}_i, \vect{a}_i) \}^{N_{\batch}}_{i=1}$ }
		\State{Update the target networks: $\vw_{\textrm{target}} \leftarrow \tau \vw + (1 - \tau) \vw_{\textrm{target}}$, 
		$\vtheta_{\textrm{target}} \leftarrow \tau \vtheta + (1 - \tau) \vtheta_{\textrm{target}} $}
%		\State{$\vw_{\textrm{target}} \leftarrow \tau \vect{w} + (1 - \tau) \vect{w}_{\textrm{target}}$}
%		\State{$\vect{\theta}_{\textrm{target}} \leftarrow \tau \vect{\theta} + (1 - \tau) \vect{\theta}_{\textrm{target}} $ }
		\EndIf
		\EndFor
		\Until{the convergence \\} 
		\Return $\vect{\theta}$
	\end{algorithmic}
	\label{alg:adInfoHRL}
\end{algorithm}

\section{Experiments}
We evaluated the proposed algorithm adInfoHRL on the OpenAI Gym platform~\citep{Brockman16} with the MuJoCo Physics
simulator~\citep{Todorov12}. 
We compared its performance with that of PPO implemented in OpenAI baselines~\citep{baselines} and TD3.
\citet{Henderson18} have recently claimed that algorithm performance varies across environment, there is thus no clearly best method for all benchmark environments, 
and off-policy and on-policy methods have advantages in different problem domains. 
To analyze the performance of adInfoHRL, we compared it with state-of-the-art algorithms for both on-policy and off-policy methods, although we focused on the comparison with TD3, 
as our implementation of adInfoHRL is based on it. 
To determine the effect of learning the latent variable via information maximization, we used the same network architectures for the actor and critic in adInfoHRL and TD3.
In addition, to evaluate the benefit of the advantage-weighted importance, we evaluated a variant of adInfoHRL, which does not use the advantage-weighted importance for computing mutual information.　We refer to this variant of adInfoHRL as infoHRL.
The gating policy updated variable $o$ once every three time steps. 
We tested the performance of adInfoHRL with two and four options.

\begin{figure}[]
	\centering
	\includegraphics[width=0.9\columnwidth]{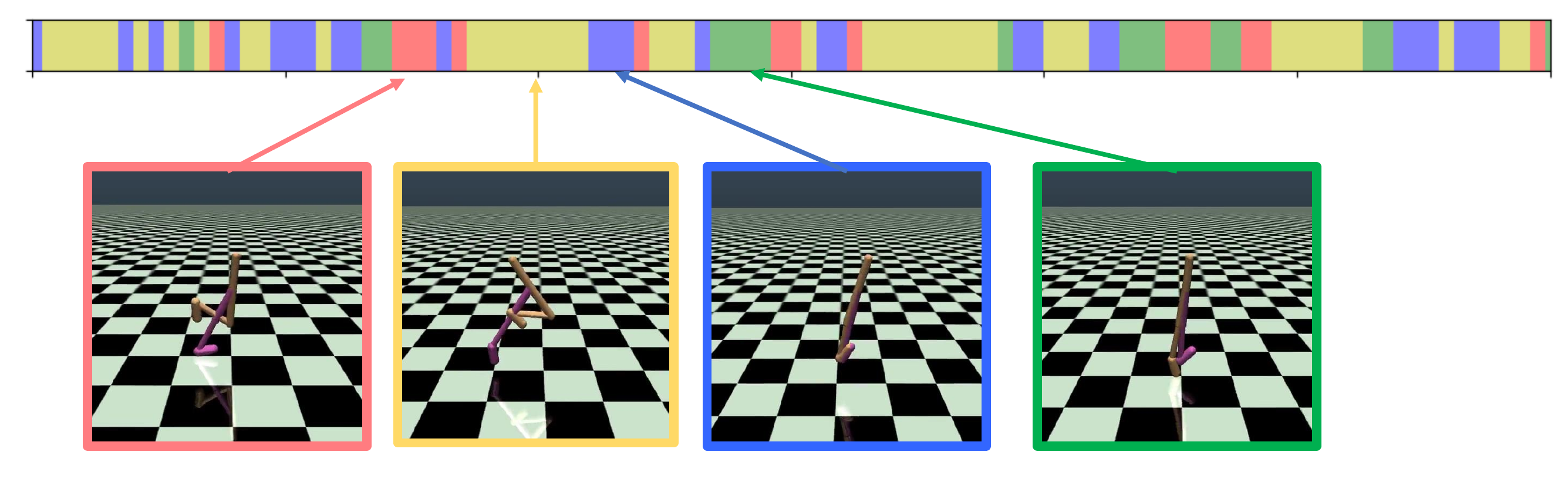}
	\caption{Activation of the four options over time steps on the Walker2d task. }
	\label{fig:option_time}
\end{figure}

The activation of options over time and snapshots of the learned option policies on the Walker2d task are shown in Figure~\ref{fig:option_time}, which visualizes the result from adInfoHRL with four options.
One can see that the option policies are activated in different phases of locomotion.
While the option indicated by yellow in Figure~~\ref{fig:option_time} corresponds to the phase for kicking the floor, 
the option indicated by blue  corresponds to the phase when the agent was on the fly.
Visualization of the options learned on the HalfCheetah and Ant tasks are shown in Appendix~\ref{app:exp_result}.

\begin{figure}[]
	\centering
	\begin{subfigure}[b]{0.32\columnwidth}
		\includegraphics[width=\textwidth]{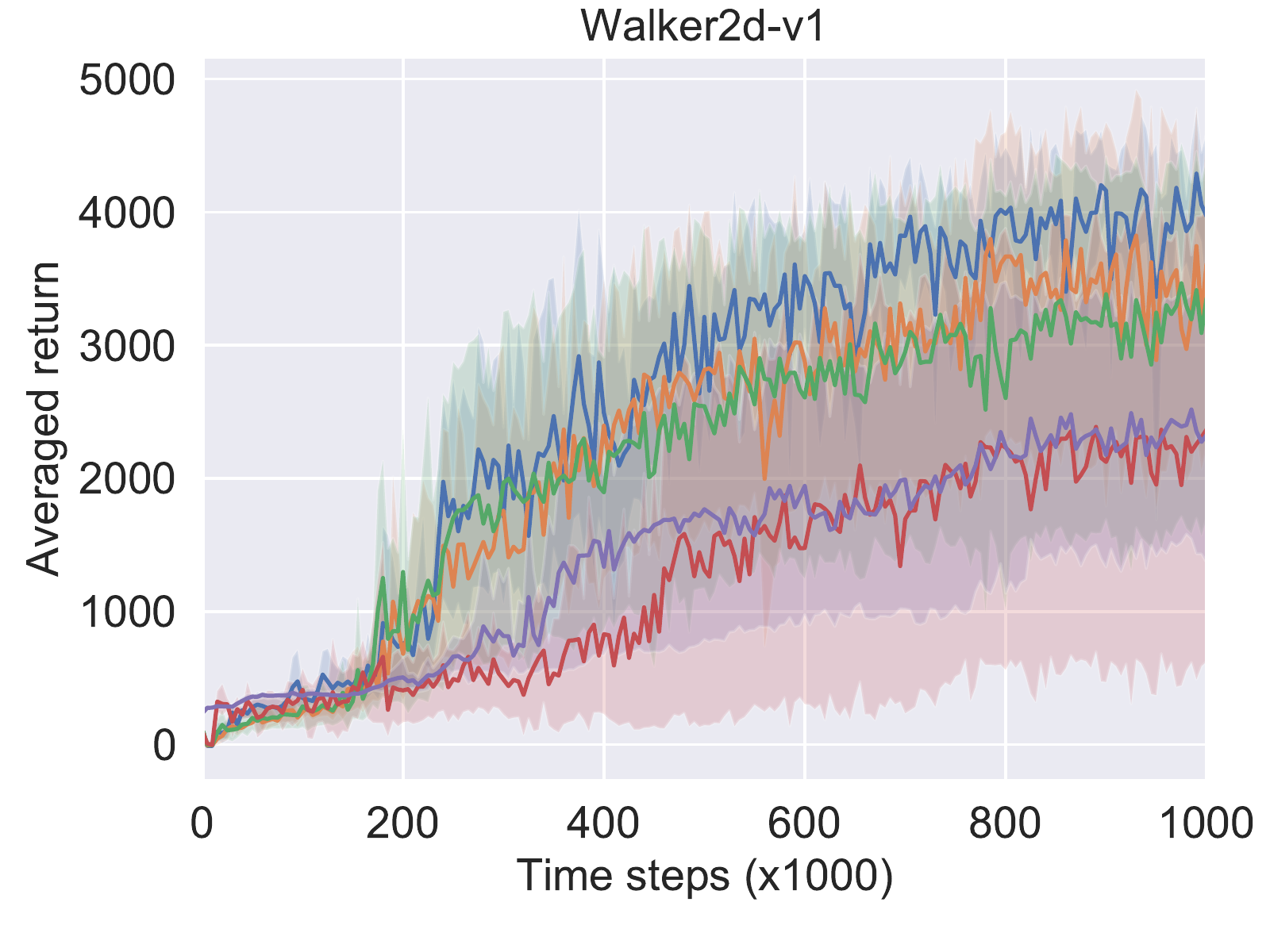}
		\caption{Results on Walker2d.}
	\end{subfigure}
	\begin{subfigure}[b]{0.32\columnwidth}
		\includegraphics[width=\textwidth]{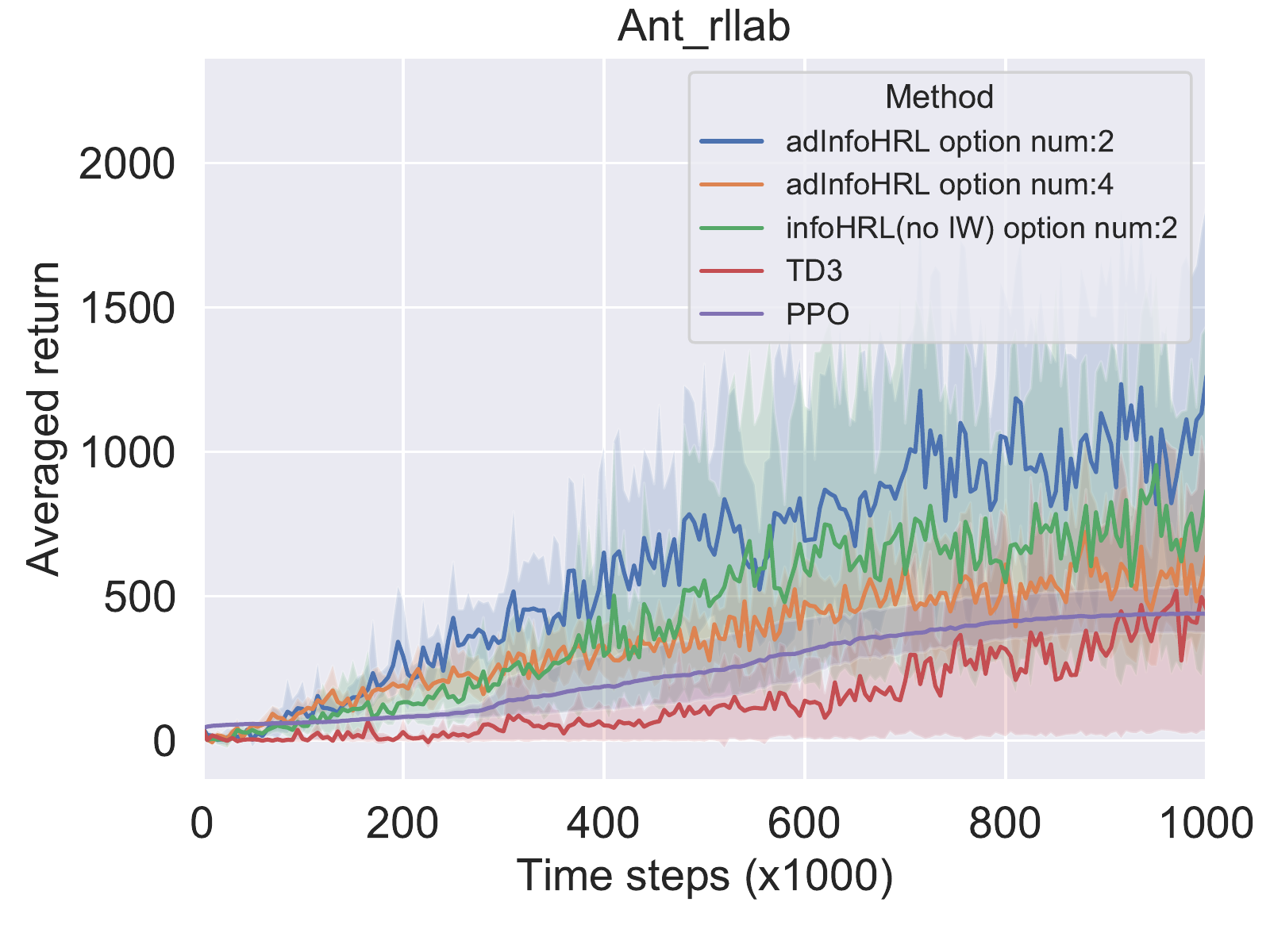}
		\caption{Results on Ant.}
	\end{subfigure}
	\begin{subfigure}[b]{0.32\columnwidth}
		\includegraphics[width=\textwidth]{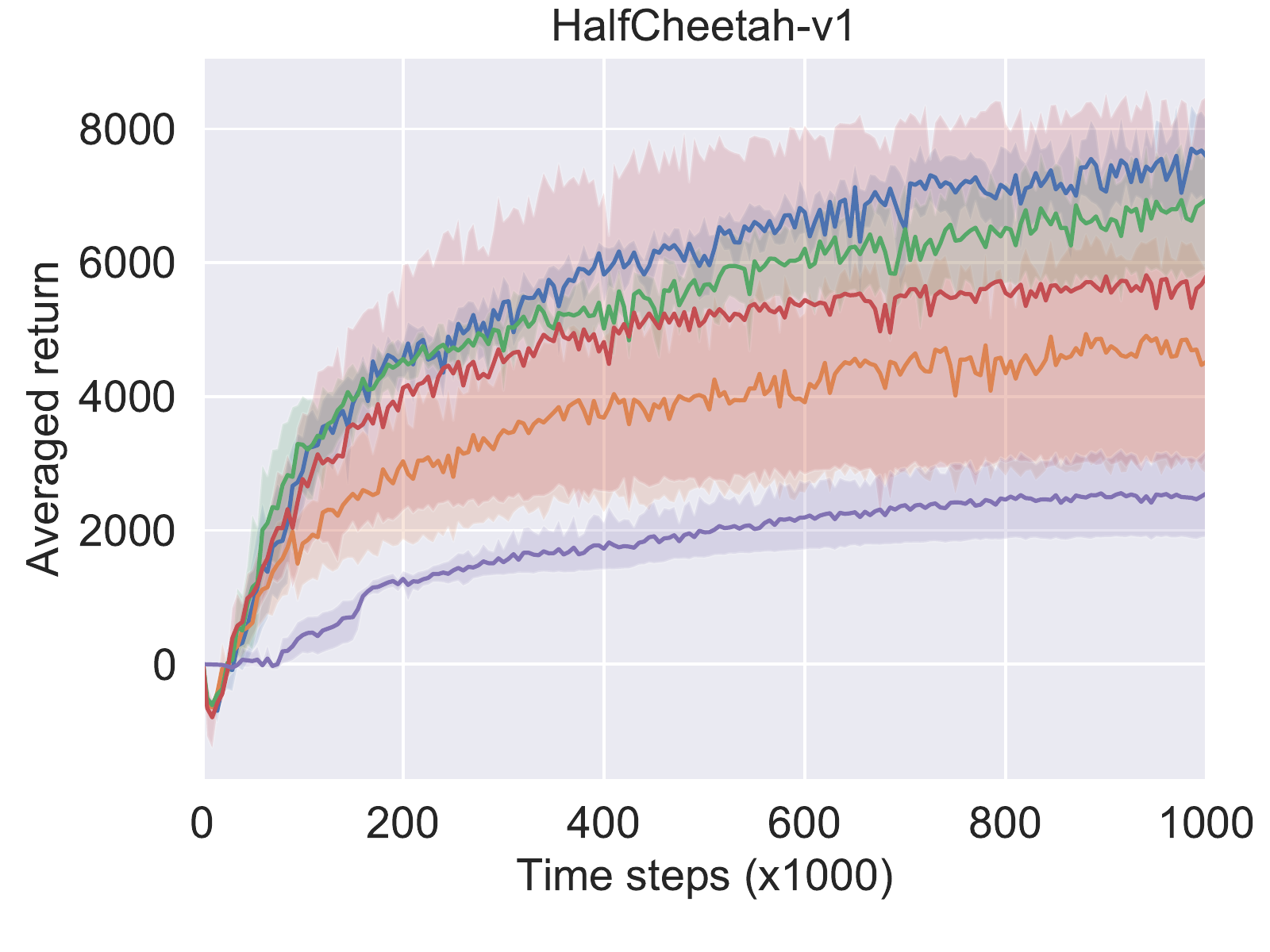}
		\caption{Results on HalfCheetah.}
	\end{subfigure}
	\begin{subfigure}[t]{0.32\columnwidth}
		\includegraphics[width=\textwidth]{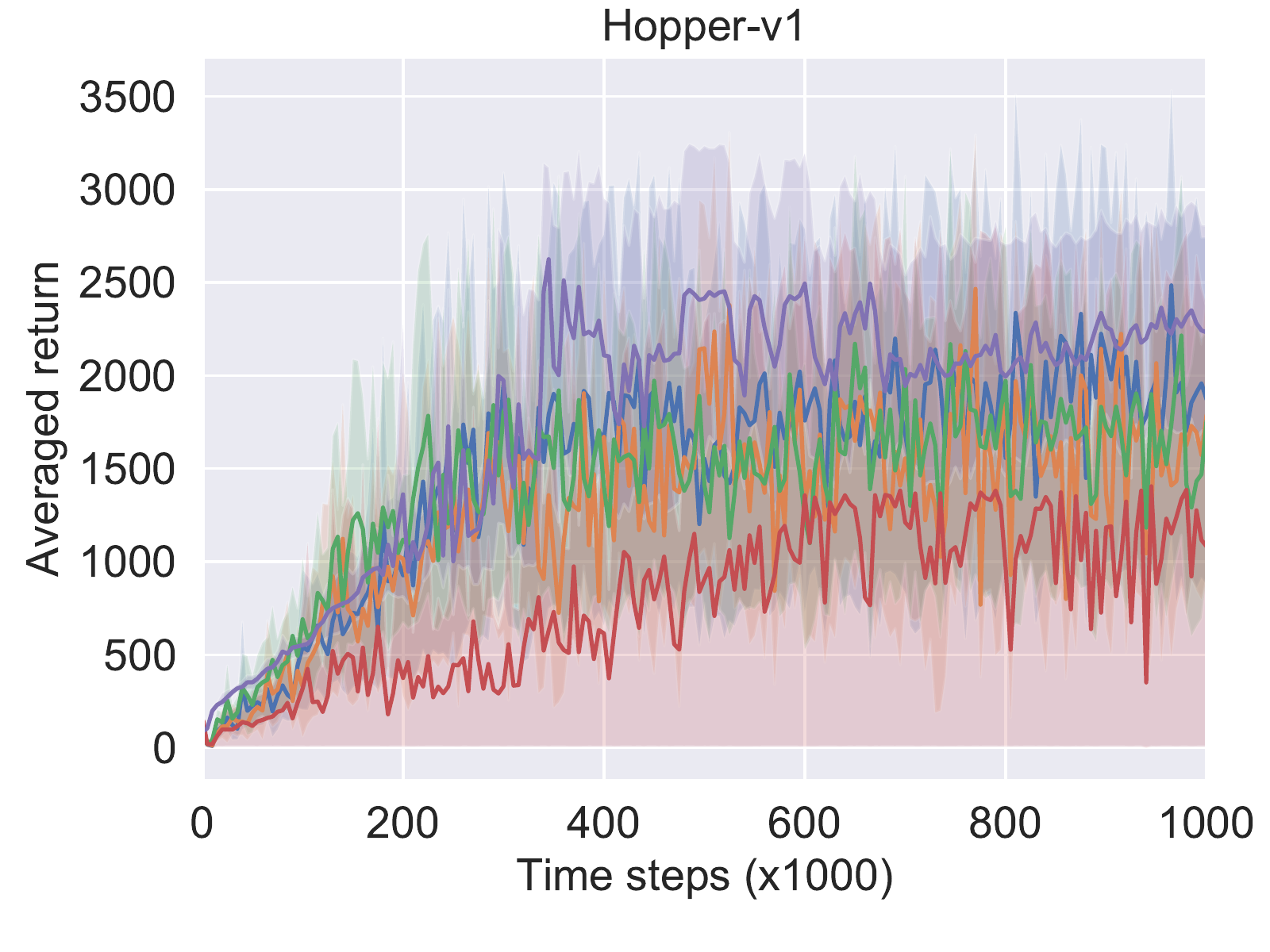}
		\caption{Results on Hopper.}
	\end{subfigure}
	\begin{subfigure}[t]{0.32\columnwidth}
		\includegraphics[width=\textwidth]{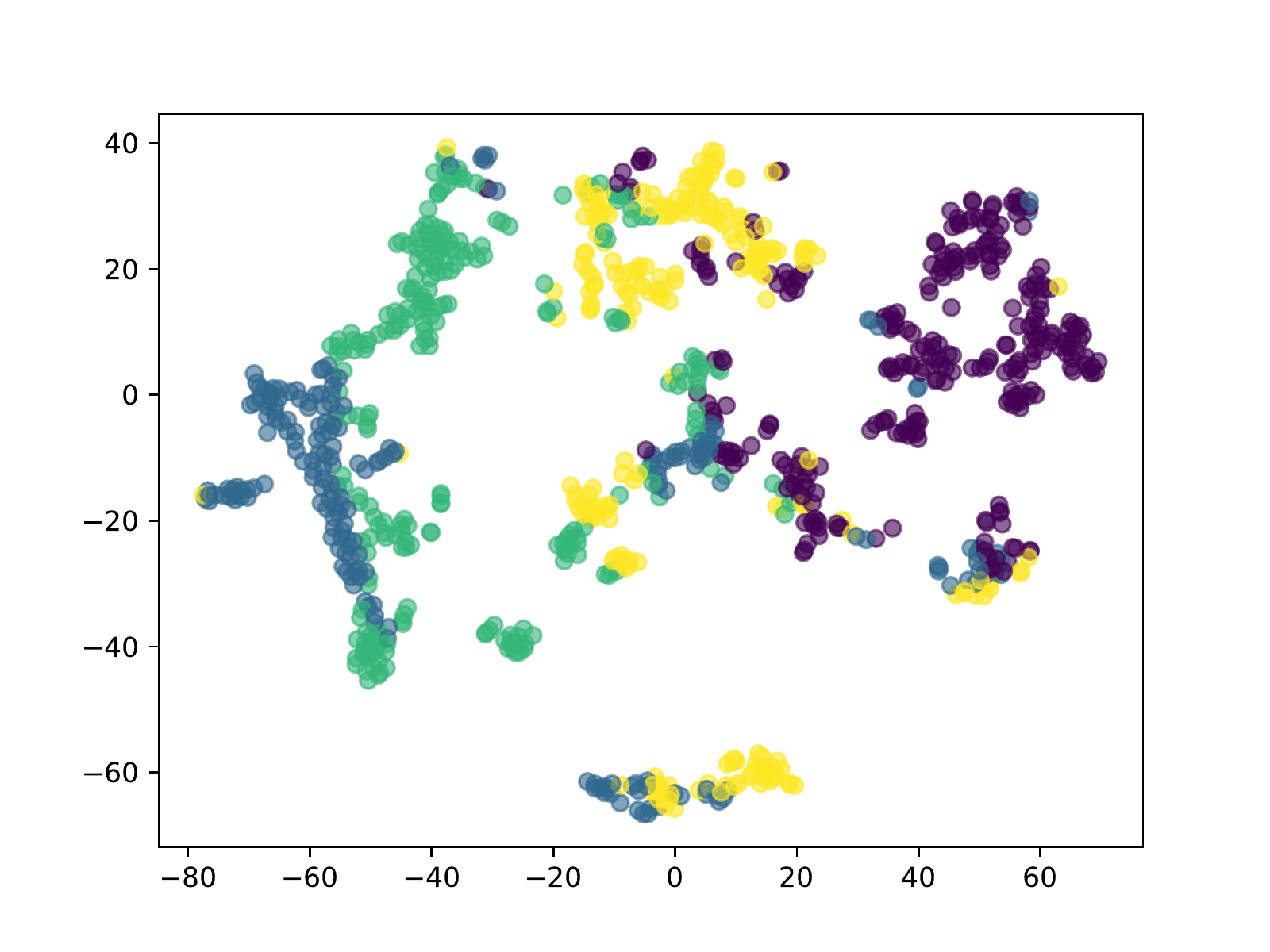}
		\caption{Output of the option network in the state-action space on Walker2d. }
	\end{subfigure}
	\begin{subfigure}[t]{0.32\columnwidth}
		\includegraphics[width=\textwidth]{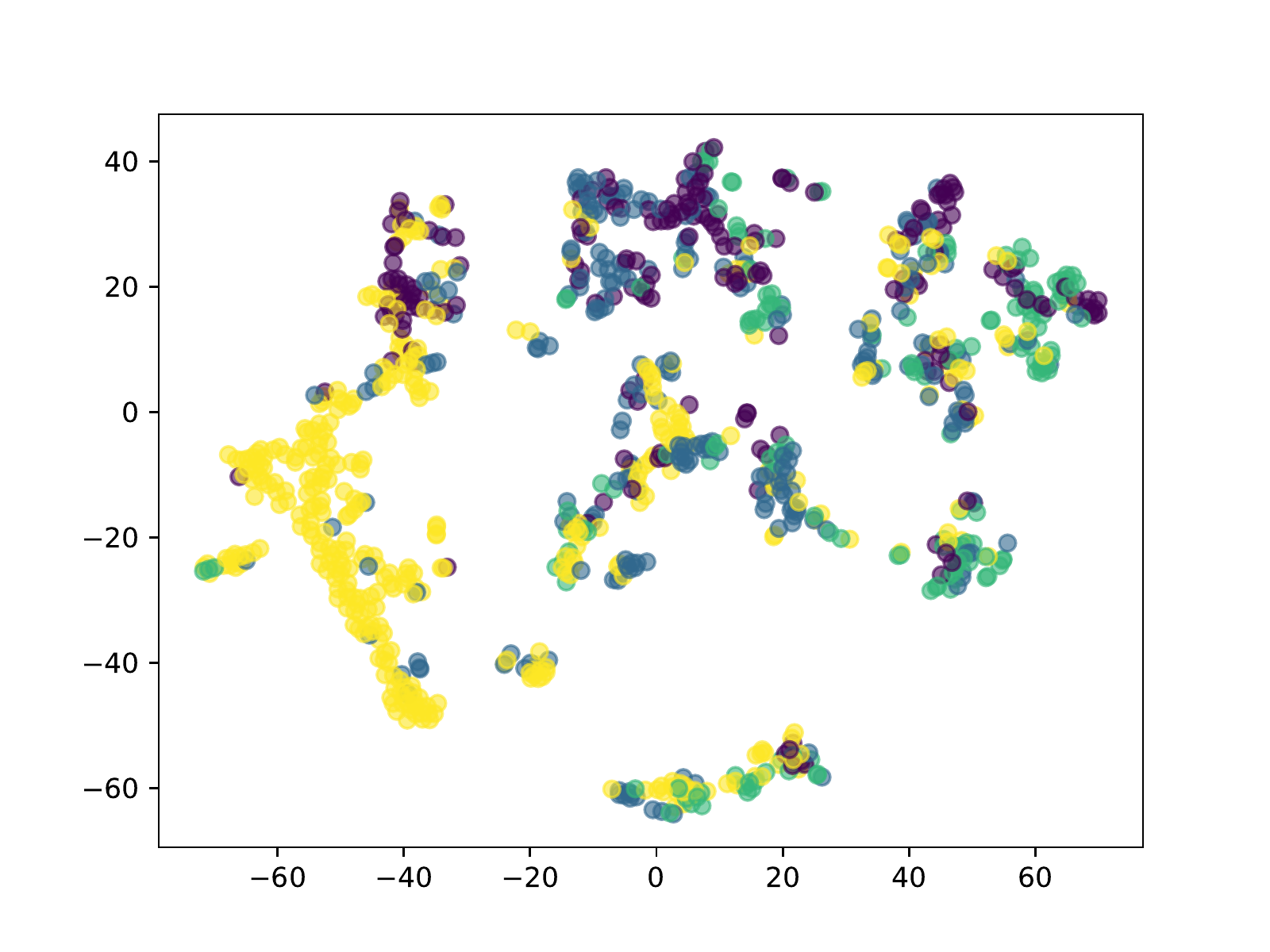}
		\caption{Activation of options in the state space on Walker2d. }
	\end{subfigure}
	\caption{Performance of adInfoHRL. (a)-(d) show comparison with baseline methods. 
		(e) and (f) show the output of the option network and the activation of options on Walker2d, respectively.
		%In (e) and (f), the dimension is reduced by T-SNE for visualization.
	}
	\label{fig:performance}
\end{figure}

The averaged return of five trials is reported in Figure~\ref{fig:performance}(a)-(d).
AdIfoHRL yields the best performance on Ant\footnote{We report the result on the Ant task implemented in rllab~\citep{Duan16} instead of Ant-v1 implemented in the OpenAI gym, since the Ant task in the rllab is known to be harder than the Ant-v1 in the OpenAI gym. Results on Ant-v1 in the OpenAI gym is reported in Appendix~\ref{app:exp_result}.} 
and Walker2d, whereas the performance of TD3 and adInfoHRL was comparable on HalfCheetah and Hopper, and PPO outperformed the other methods on Hopper.
\citet{Henderson18} claimed that on-policy methods show their superiority on tasks with unstable dynamics, 
and our experimental results are in line with such previous studies.
%The results show that the optimal number of options is dependent on the task: two options for HalfCheetah and Ant, and four options for Walker2d.
AdinfoHRL outperformed infoHRL, which is　the variant of adInfoHRL without the advantage-weighted importance on all the tasks. 
This result shows that the adavatage-weighted importance enhanced the performance of learning options.

AdInfoHRL exhibited the sample efficiency on Ant and Walker2d in the sense that it required fewer samples than TD3 to achieve comparable performance on those tasks.
%The performance of adInfoHRL with 10 options is lower than TD3 on all the tasks, which indicates that too many options can deteriorate the learning performance.
The concept underlying adInfoHRL is to divide the state-action space to deal with the multi-modal advantage function and learn option policies corresponding to separate modes of the advantage function.
Therefore, adInfoHRL shows its superiority on tasks with the multi-modal advantage function and not on tasks with a simple advantage function. 
Thus, it is natural that the benefit of adInfoHRL is dependent on the characteristics of the task.

The outputs of the option network and the activation of options on Walker2d are shown in Figure~\ref{fig:performance}(e)-(f), which visualize the result from adInfoHRL with four options. 
For visualization, the dimensionality was reduced using t-SNE~\citep{Maaten08}.
The state-action space is clearly divided into separate domains in Figure~\ref{fig:performance}(e).
As shown in Figure~\ref{fig:performance}(f), the options are activated in different domains of the state space, which indicates that diverse options are learned by adInfoHRL.

%The intuition behind our approach is that there may exist multiple actions that can achieve comparably high returns under the same state.
%\begin{figure}[]
%	\centering
%	\begin{subfigure}[b]{0.34\columnwidth}
%		\includegraphics[width=\textwidth]{clustering_Ant-v1}
%		\caption{Output of the option network in the state-action space on Ant-v1 task. The dimension is reduced with T-SNE.}
%	\end{subfigure}
%	\begin{subfigure}[b]{0.34\columnwidth}
%		\includegraphics[width=\textwidth]{activation_Ant-v1}
%		\caption{Activation of options in the state space on Ant-v1 task. For visualization, the dimension is reduced with T-SNE.}
%	\end{subfigure}
%	\caption{Activation of the options.}
%	\label{fig:activation}
%\end{figure}

%In the second experiment, we trained a policy on a Ant task based on rllab, where the agent receives a reward no matter which direction it goes.
%On this task, TD3 does not work at all.
%In contrast, adInfoHRL showed the better learning performance as shown in Figure~\ref{fig:AntMaze_nowall}(a). 
%We visualized the behavior of learned options on this Ant task in Figure~\ref{fig:AntMaze_nowall}(b).
%One can see that each option learned different behaviors to achieve the task.
%This result indicates that adInfoHRL can learn diverse option policies through maximizing the mutual information. 
%
%\begin{figure}[]
%	\centering
%	\includegraphics[width=0.33\columnwidth]{AntMaze_nowall}
%	\caption{Performance on the AntMaze without walls.}
%	\label{fig:AntMaze_nowall}
%\end{figure}

\section{Related Work and Discussion}
%Our policy network structure is similar to the ones in~\cite{Vezhnevets17,Florensa17} in the sense that 
%a lower level network receives a latent variable generated by a higher level network as a part of an input.
%However, there is no clear rationale to obtain meaningful representations in those studies.

%HRL has attracted great attention in recent years, and recent studies addressed how to learn the latent variable in HRL.
Past studies have proposed several ways to deal with the latent variable in HRL.
The recent work by~\citet{Smith18} proposed inferred option policy gradients~(IOPG), which is derived as an extension of policy gradient to the option framework.
\citet{Nachum18} recently proposed off-policy target correction for HRL on goal-oriented tasks,
where a higher-level policy instructs a lower-level policy by generating the goal signal instead of an inferred latent variable.
A popular approach for learning the latent variable in HRL is the variational approach.
The recent work by~\citet{Haarnoja18} is based on soft actor critic~\citep{Haarnoja18b}, and the latent variable is inferred using the variational approach.
The work by \citet{Hausman18} is also closely related to the variational approach, 
and they proposed a method for learning a latent variable of a hierarchical policy via a variational bound.
%They derived an off-policy HRL algorithm based on stochastic value gradients. 
On the contrary, our method learns the latent variable by maximizing MI with advantage-weighted importance.
Recent studies by \citet{Gregor16,Florensa17,Eysenbach18} also considered the MI in their formulation.
In these methods, MI between the state and the latent variable is considered so as to obtain diverse behaviors.
Our approach is different from the previous studies in the sense that we employ MI between the latent variable and the state-action pairs, which leads to the division of the state-action space instead of considering only the state space.
We think that dividing the state-action space is an efficient approach when the advantage function is multi-modal, as depicted in Figure~\ref{fig:schematic}.
InfoGAIL proposed by~\citet{Li17} learns the interpretable representation of the state-action space via MI maximization.
InfoGAIL can be interpreted as a method that divides the state-action space based on the density induced by an expert's policy 
by maximizing the regularized MI objective.
In this sense, it is closely related to our method, although their problem setting is imitation learning~\citep{Osa18c}, which is different from our HRL problem setting.

The use of the importance weight based on the value function has appeared in previous studies~\citep{Dayan97,Kober11,Neumann09,Osa18}. 
For example, the method proposed by \citet{Neumann09} employs the importance weight based on the advantage function for learning a monolithic policy, while our method uses a similar importance weight for learning a latent variable of a hierarchical policy.
Although \citet{Osa18} proposed to learn a latent variable in HRL with importance sampling, their method is limited to episodic settings where only a single option is used in an episode.

Our method can be interpreted as an approach that divides the state-action space based on the MI criterion. 
This concept is related to that of Divide~and~Conquer~(DnC) proposed by~\citet{Ghosh18}, although DnC clusters the initial states and does not consider switching between option policies during the execution of a single trajectory.

%A recent study by \citet{Rajeswaran17} indicates that a relatively simple policy architecture can achieve high performance in continuous control tasks.
%The results of this study do not conflict with that indication. 
%The main conclusion from this work is that learning the latent discrete representation of the state-action space can be beneficial for learning,
%and we think that it can be achieved with simple policy models.
%Recent HRL methods such as the FeUdal Network~\cite{Vezhnevets17} and Option Critic~\cite{Bacon17} are on-policy.
%However, as indicated in~\cite{Henderson18}, on-policy and off-policy approaches show their superiority in different domains,
%and we therefore think that it is beneficial to develop off-policy HRL methods. 
%Since our method is applicable to any problem where off-policy RL methods are usable, 
%our approach can be combined with recent techniques such as Hindsight Experience Replay~(HER) proposed in~\cite{Andrychowicz17}.
%In contrast, since existing methods such as FeUdal network and Option Critic are on-policy, they cannot be combined with HER.

%There are several ways to potentially improve adInfoHRL presented in this work.
%Recent work shows that combining various techniques for Q-learning can enhance the performance drastically~\cite{Hessel18}.
%Since the our formulation of learning the gating policy $\pi(o|\vs)$ is analogous to the standard Q-learning,
%our method will also benefit from combining existing various techniques for Q-learning.
In this study we developed adInfoHRL based on deterministic option policies.
However, the concept of dividing the state-action space via advantage-weighted importance can be applied to stochastic policy gradients as well.
Further investigation in this direction is necessary in future work.

\section{Conclusions}
We proposed a novel HRL method, hierarchical reinforcement learning via advantage-weighted information maximization.
In our framework, the latent variable of a hierarchical policy is learned as a discrete latent representation of the state-action space.
Our HRL framework is derived by considering deterministic option policies and by leveraging the analogy between the gating policy for HRL and a monolithic policy for the standard RL.
%In infoHRL, the gating policy is formulated based on MaxEnt RL to encode the exploration over options in its policy form,
%and the option policy is trained based on deterministic policy gradient.
The results of the experiments indicate that adInfoHRL can learn diverse options on continuous control tasks.
Our results also suggested that our approach can improve the performance of TD3 in certain problem domains.
%Although we instantiated infoHRL based on DPG and soft Q-learning, one can extend this framework based on other policy-gradient-based methods and variants of Q-learning. 
%It is necessary to investigate how to improve the framework of infoHRL in future work.

%\subsection{Tables}
%
%All tables must be centered, neat, clean and legible. Do not use hand-drawn
%tables. The table number and title always appear before the table. See
%Table~\ref{sample-table}.
%
%Place one line space before the table title, one line space after the table
%title, and one line space after the table. The table title must be lower case
%(except for first word and proper nouns); tables are numbered consecutively.
%
%\begin{table}[t]
%\caption{Sample table title}
%\label{sample-table}
%\begin{center}
%\begin{tabular}{ll}
%\multicolumn{1}{c}{\bf PART}  &\multicolumn{1}{c}{\bf DESCRIPTION}
%\\ \hline \\
%Dendrite         &Input terminal \\
%Axon             &Output terminal \\
%Soma             &Cell body (contains cell nucleus) \\
%\end{tabular}
%\end{center}
%\end{table}
%

\subsubsection*{Acknowledgments}
MS was partially supported by KAKENHI 17H00757.
%
%Use unnumbered third level headings for the acknowledgments. All
%acknowledgments, including those to funding agencies, go at the end of the paper.

\bibliography{iclr19}
\bibliographystyle{iclr2019_conference}

\newpage

\appendix

\section{Mutual Information with Advantage-Weighted Importance}
%\subsection{Mutual Information with Advantage-Weighted Importance}
\label{app:weighted_info}
%Prior work~\cite{Gomes10,Hu17} employed the empirical estimate of the mutual information given by $\hat{I}_{\vect{w}}(y; \vx) = \hat{H}(y)  - \hat{H}(y|\vx)$
%where $\hat{H}(y) = - \hat{p}(y; \vw) \log \hat{p}(y; \vw)$ and $\hat{p}(y; \vw)$ is 
%the empirical estimate of $p(y)= \int p(\vx) p(y | \vx) \intd \vx = \E[ p(y | \vx) ]$ given by
%\begin{align}
%\hat{p}(y; \vw) = \frac{1}{N} \sum_{i=1}^{N} \hat{p}(y_i | \vx_i; \vw).
%\end{align}
%Likewise, $\hat{H}(y|\vx)$ is the empirical estimates of $H(y|\vx)= - \E \left[ p( y | \vx) \log p( y | \vx) \right]$ 
%given by 
%\begin{align}
%\hat{H}(y|\vx) = - \frac{1}{N} \sum_{i}^{N} \hat{p}(y_i | \vx_i; \vect{w}) \log \hat{p}( y_i | \vx_i; \vw).
%\end{align}
The mutual information~(MI) between the latent variable $o$ and the state action pair $(\vs, \va)$ is defined as
\begin{align}
I\big((\vs, \va), o\big) = H(o) - H(o | \vs, \va)
\end{align}
where $H(o) = \int p(o) \log p(o) \intd o$ and $H(o|\vs, \va) = \int p(o|\vs, \va) \log p(o|\vs, \va) \intd o$.
We make the empirical estimate of MI employed by \citet{Gomes10,Hu17} and modify it to employ the importance weight.
The empirical estimate of MI with respect to the density induced by a policy $\pi$ is given by
\begin{align}
\hat{I}(\vs, \va; o) = \sum_{o \in \mathcal{O}} \hat{p}(o) \log \hat{p}(o) - \hat{H}(o|\vs, \va).
\label{eq:emp_I}
\end{align}
%We employ empirical estimate of the mutual information
%\begin{align}
%I(o; \vect{s}, \vect{a}) = H\{ \hat{p}(o; \vect{w}) \} - \frac{1}{N} \sum_{i=1}^{N} H\{ p(o | \vect{s}_i, \vect{a}_i; \vect{w}) \} 
%\end{align}
%where empirical estimate of $p(o)$ can be obtained by
%\begin{align}
%\hat{p}(o) = \int p(\vect{s}, \vect{a}) p(o | \vect{s}, \vect{a}) \intd\vect{a}\intd\vect{s}\approx \frac{1}{N} \sum_{i=1}^{N} p(o | \vect{s}_i, \vect{a}_i) 
%\end{align}
We consider the case where we have samples collected by a behavior policy $\beta(\vs|\va)$ and need to estimate MI with respect to the density induced by policy $\pi$.
Given a model $p(o | \vs, \va; \vect{\eta})$ parameterized by vector $\vect{\eta}$, 
$p(o)$ can be rewritten as
\begin{align}
p(o) = \int p^{\beta}(\vect{s}, \vect{a}) \frac{p^{\pi}(\vs, \va)}{p^{\beta}(\vs, \va)} p(o | \vect{s}, \vect{a}; \vect{\eta}) \intd\vect{a}\intd\vect{s} 
= \E \left[ W(\vs, \va) p(o | \vect{s}, \vect{a}; \vect{\eta})  \right],
\end{align}
where $W(\vs, \va) =  \frac{p^{\pi}(\vs, \va)}{p^{\beta}(\vs, \va)}$ is the importance weight.
Therefore, the empirical estimate of $p(o)$ with respect to the density induced by a policy $\pi$ is given by
\begin{align}
\hat{p}(o) = \frac{1}{N} \sum_{i=1}^{N} \tilde{W}(\vs_i, \va_i) p(o | \vs_i, \va_i; \vect{\eta}),
\label{eq:emp_p}
\end{align}
where $\tilde{W}(\vect{s}, \vect{a}) = \frac{\tilde{W}(\vs, \va)}{\sum^{N}_{j=1}\tilde{W}(\vs_j, \va_j) }$ is the normalized importance weight.

Likewise, the conditional entropy with respect to the density induced by a policy $\pi$ is given by
\begin{align}
H(o|\vs, \va) & = \int p^{\pi}(\vs, \va) p(o | \vs, \va; \vect{\eta}) \log p(o|\vs, \va; \vect{\eta}) \intd\vs\intd\va \\
& = \int p^{\beta}(\vs, \va) \frac{p^{\pi}(\vs, \va)}{p^{\beta}(\vs, \va)}  p(o | \vs, \va; \vect{\eta}) \log p(o|\vs, \va; \vect{\eta}) \intd\vs\intd\va \\
& = \E \left[ W(\vs, \va) p(o | \vs, \va; \vect{\eta}) \log p(o | \vs, \va; \vect{\eta}) \right].
\end{align}
Therefore, the empirical estimate of the conditional entropy with respect to the density induced by a policy $\pi$ is given by
\begin{align}
\hat{H}(o|\vect{s}, \vect{a}) = \frac{1}{N} \sum_{i=1}^{N} W(\vect{s}_i, \vect{a}_i) p(o | \vect{s}_i, \vect{a}_i; \vect{\eta}) \log p(o | \vect{s}_i, \vect{a}_i; \vect{\eta}).
\label{eq:emp_H}
\end{align}
Thus, the empirical estimates of MI can be computed by Equations~(\ref{eq:emp_I}), (\ref{eq:emp_p}) and (\ref{eq:emp_H}). 

\section{Derivation of the state-value function}
In HRL, the value function is given by
\begin{align}
V(\vs) 
=  \int  \sum_{o  \in \mathcal{O}} \pi( o | \vect{s} ) \pi( \vect{a} | \vect{s}, o ) Q^{\pi}( \vect{s}, \vect{a} ) \textrm{d}\vect{a}
=  \sum_{o  \in \mathcal{O}} \pi( o | \vect{s} ) \int \pi( \vect{a} | \vect{s}, o ) Q^{\pi}( \vect{s}, \vect{a} ) \textrm{d}\vect{a}
\end{align}
Since option policies are deterministic given by $\vmu^o_{\vtheta}(\vect{s})$, the state-value function is given by
\begin{align}
V(\vs) 
=  \sum_{o  \in \mathcal{O}} \pi( o | \vs ) Q^{\pi}( \vs, \vmu^o_{\vtheta}(\vect{s}) ) \textrm{d}\va.
\end{align}

%\begin{align}
%\tilde{W}( \vs, \va)  = \frac{W( \vs, \va)}{ \sum_{j=1}^{N} W( \vs_j, \va_j) }  
%= \frac{ \frac{f(A(\vs, \va) )}{Z \beta(\va | \vs)} }{ \sum_{j=1}^{N} \frac{f(A(\vs_j, \va_j) )}{Z \beta(\va_j | \vs_j)}} 
%= \frac{ \frac{f(A(\vs, \va) )}{\beta(\va | \vs)} }{ \sum_{j=1}^{N} \frac{f(A(\vs_j, \va_j) )}{ \beta(\va_j | \vs_j)}}.
%\end{align}

%\section{Gating policy}
%We consider the gating policy given in Equation~(\ref{eq:gate}):
%\begin{align}
%\pi(o | \vs) % = \softmax\limits_{o} Q^{\pi}(\vect{s}, \vect{\mu}_{\vtheta}}(\vect{s}, o)) 
%= \frac{\exp\big(Q^{\pi}(\vs, \vmu^{o}_{\vtheta}(\vs); \vw ) \big)}{\sum_{o \in \mathcal{O}} \exp\left(Q^{\pi}\big(\vs, \vmu^{o}_{\vtheta}(\vs) ; \vw \big) \right) }.
%\nonumber
%\end{align}
%In our study the hierarchical policy is of the form
%\begin{align}
%\pi(\vect{a}|\vect{s}) = \sum_{o \in \mathcal{O}} \pi(\vect{o}|\vect{s}) \pi(\vect{a}|\vect{s}, o),
%\nonumber
%\end{align}
%and option policies are deterministic: $\pi(\vect{a}|\vect{s}, o) = \vmu^o_{\vtheta}(\vect{s})$.
%By combining these equations, we can obtain
%\begin{align}
%\pi(\vect{a}|\vect{s}) & = \sum_{o \in \mathcal{O}} \pi(\vect{o}|\vect{s}) \pi(\vect{a}|\vect{s}, o)\\
%& = \sum_{o \in \mathcal{O}} \pi(\vect{o}|\vect{s}) \vmu^o_{\vtheta}(\vect{s})
%\end{align}

\section{Experimental Details}
\label{app:exp}
We performed evaluations using benchmark tasks in the OpenAI Gym platform~\citep{Brockman16} with Mujoco physics simulator~\citep{Todorov12}. 
Hyperparameters of  reinforcement learning methods used in the experiment are shown in Tables~\ref{tbl:param_adInfoHRL}-\ref{tbl:param_PPO}.
For exploration, both adInfoHRL and TD3 used the clipped noise drawn from the normal distribution as $\epsilon \sim \textrm{clip} \big(\mathcal{N}(0, \sigma), -c, c \big)$,
where $\sigma=0.2$ and $c=0.5$.
For hyperparameters of PPO, we used the default values in OpenAI baselines~\citep{baselines}. 
For the Walker2d, HalfCheetah, and Hopper tasks, we used the Walker2d-v1, HalfCHeetah-v1, and Hopper-v1 in the OpenAI Gym, respectively.
For the Ant task, we used the AntEnv implemented in the rllab~\citep{Duan16}.
When training a policy with AdInfoHRL, infoHRL, and TD3, critics are trained once per time step, and actors are trained once every after two updates of the critics.
The source code is available at \url{https://github.com/TakaOsa/adInfoHRL}.

We performed the experiments five times with different seeds, and reported the averaged test return where the test return was computed once every 5000 time steps 
by executing 10 episodes without exploration.
When performing the learned policy without exploration, the option was drawn as
\begin{align}
o = \max_{o'} Q^{\pi}( \vs, \vmu^{o'}(\vect{s}) ),
\end{align}
instead of following the stochastic gating policy in Equations~(\ref{eq:gate}).

%The environment used in the second experiment in Section~\ref{sec:exp} is based on AntMaze in rllab, and we removed all the wall from the AntMaze task.
%We set the goal reward to zero, and the reward from the wrapped environment to 1.

\begin{table}[]
	\caption{Hyperparameters of adInfoHRL used in the experiment.}
	\centering
	\begin{tabular}{lcl}
		\hline
		%\multicolumn{2}{c}{Part}                   \\
		%\cmidrule{1-2}
		Description     & Symbol     & Value \\
		\hline
		Coefficient for updating the target network & $\tau$  & 0.005     \\
		Discount factor     & $\gamma$ & 0.99      \\
		Learning rate for actor & & 0.001 \\
		Learning rate for critic & & 0.001 \\
		Batch size for critic & & 100 \\
		Total batch size for all option policies & & 200 (option num=2), 400 (option num=4) \\
		Batch size for the option network & & 50 \\
		Size of the on-policy buffer & & 5000 \\
		Number of epochs for training the option network & & 40 \\
		Number of units in hidden layers     &        & (400, 300)  \\
		Activation function     &        & Relu, Relu, tanh  \\
		optimizer & & Adam\\
		noise clip threshold & c & 0.5 \\
		noise for exploration &  & 0.1 \\
		action noise for the critic update&  & 0.2 \\
		variance of the noise for MI regularization &  & 0.04 \\
		coefficient for the MI term& $\lambda$ & 0.1 \\ 
		\hline
	\end{tabular}
	\label{tbl:param_adInfoHRL}
\end{table}

\begin{table}[]
	\caption{Hyperparameters of TD3 used in the experiment.}
	\centering
	\begin{tabular}{lcl}
		\hline
		%\multicolumn{2}{c}{Part}                   \\
		%\cmidrule{1-2}
		Description     & Symbol     & Value \\
		\hline
		Coefficient for updating the target network & $\tau$  & 0.005     \\
		Discount factor     & $\gamma$ & 0.99      \\
		Learning rate for actor & & 0.001 \\
		Learning rate for critic & & 0.001 \\
		Batch size & & 100 \\
		Number of units in hidden layers     &        & (400, 300)  \\
		Activation function     &        & Relu, Relu, tanh  \\
		optimizer & & Adam\\
		noise clip threshold & c & 0.5 \\
		noise for exploration &  & 0.1 \\
		action noise for the critic update&  & 0.2 \\
		\hline
	\end{tabular}
	\label{tbl:param_td3}
\end{table}

\begin{table}[]
	\caption{Hyperparameters of PPO used in the experiment. We tuned hyperparameters for our tasks, which are defferent from the default parameters in OpenAI baselines~\citep{baselines}. }
	\centering
	\begin{tabular}{lll}
		\hline
		%\multicolumn{2}{c}{Part}                   \\
		%\cmidrule{1-2}
		Description     & Symbol     & Value \\
		\hline
		Coefficient for updating the target network & $\tau$  & 0.001     \\
		Discount factor     & $\gamma$ & 0.99      \\
		Batch size & & 2048 \\
		Number of units in hidden layers     &        & (64, 64)  \\
		Clipping parameter& $\epsilon$  & 0.15 \\
		Initial learning rate &   & 0.0005 \\
		Learning rate schedule  &   & linear \\
		\hline
	\end{tabular}
	\label{tbl:param_PPO}
\end{table}

\newpage

\section{Additional Information on Experimental Results }
\label{app:exp_result}

\begin{figure}[]
	\centering
	\begin{subfigure}[t]{0.4\columnwidth}
		\includegraphics[width=0.8\textwidth]{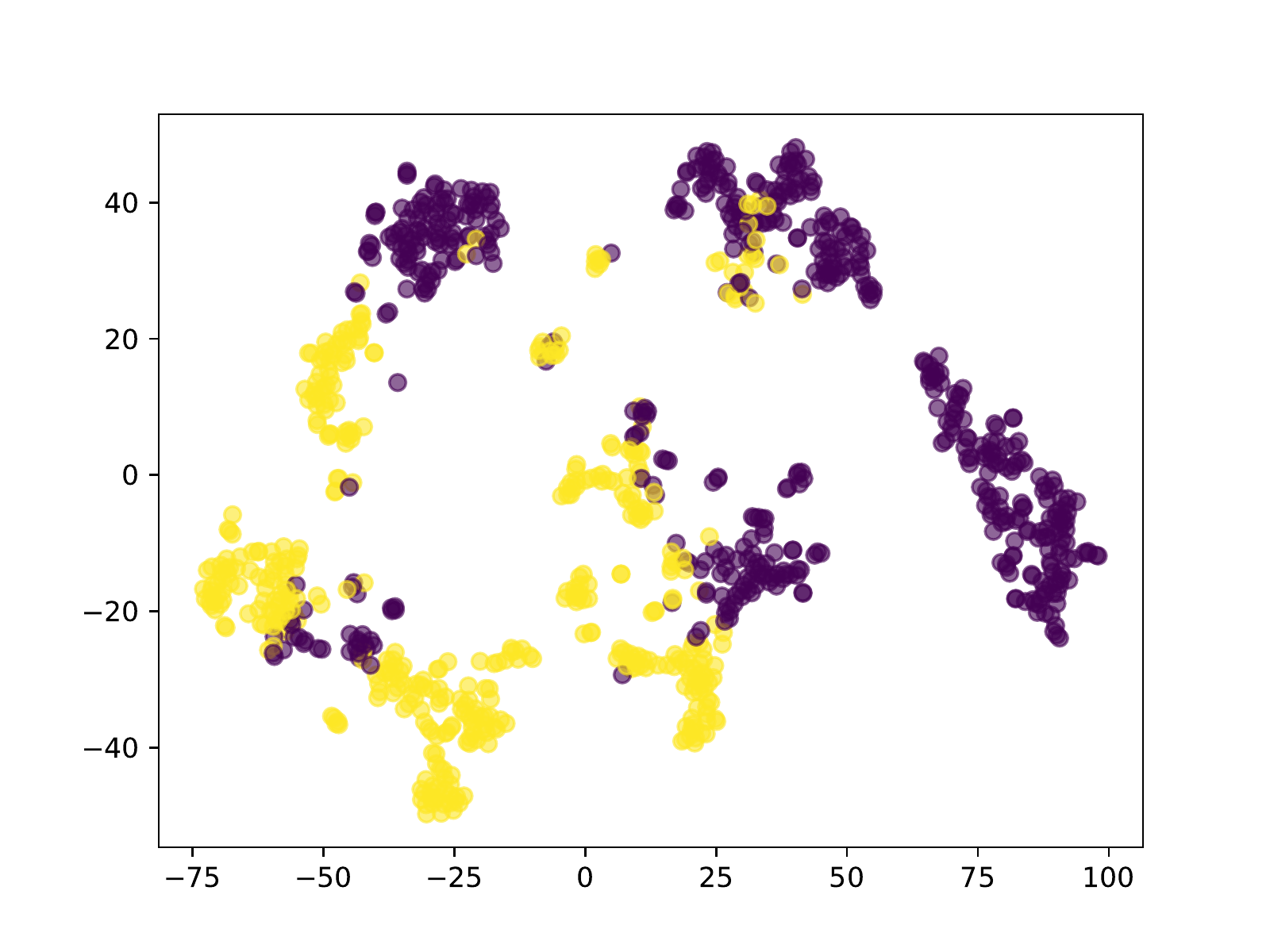}
		\caption{Output of the option network in the state-action space on HalfCheetah-v1. }
	\end{subfigure}
	\begin{subfigure}[t]{0.4\columnwidth}
		\includegraphics[width=0.8\textwidth]{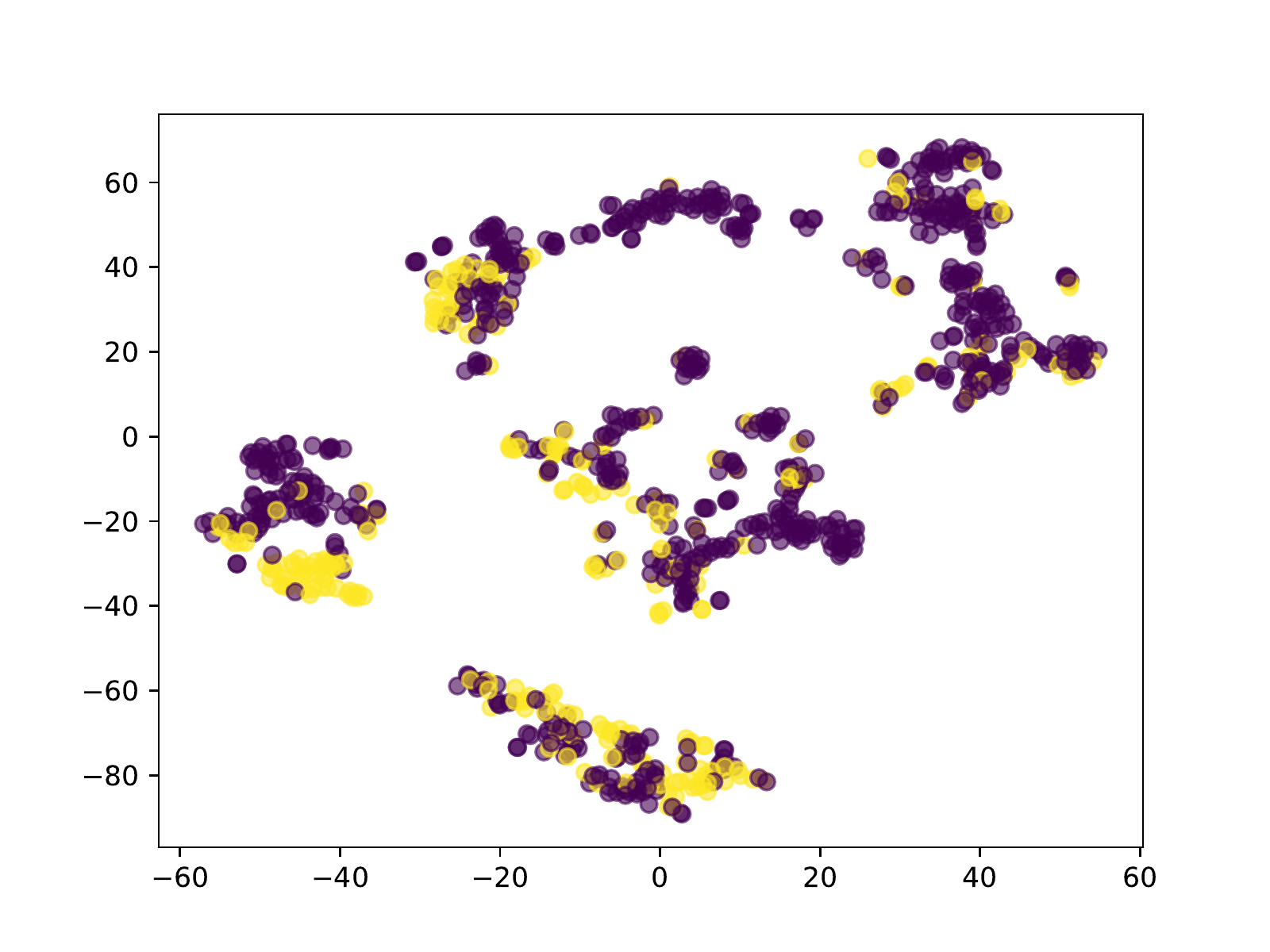}
		\caption{Activation of options in the state space on HalfCheetah-v1. }
	\end{subfigure}
	\caption{Distribution of options on the HalfCheetah-v1 task using adInfoHRL with two options. The dimensionality is reduced by t-SNE for visualization.}
	\label{fig:performance_HalfCheetah}
\end{figure}

On the HalfCheetah task, adInfoHRL delivered the best performance with two options.
The distribution of options on HalfCheetah0v1 after one million steps is shown in Figure~\ref{fig:performance_HalfCheetah}.
Although the state-action space is evenly divided, the options are not evenly activated.
This behavior can occur because the state-action space is divided based on the density induced by the behavior policy
while the activation of options is determined based on the quality of the option policies in a given state.
Moreover, an even division in the action-state space is not necessarily the even division in the state space.

\begin{figure}[]
	\centering
	\includegraphics[width=0.9\columnwidth]{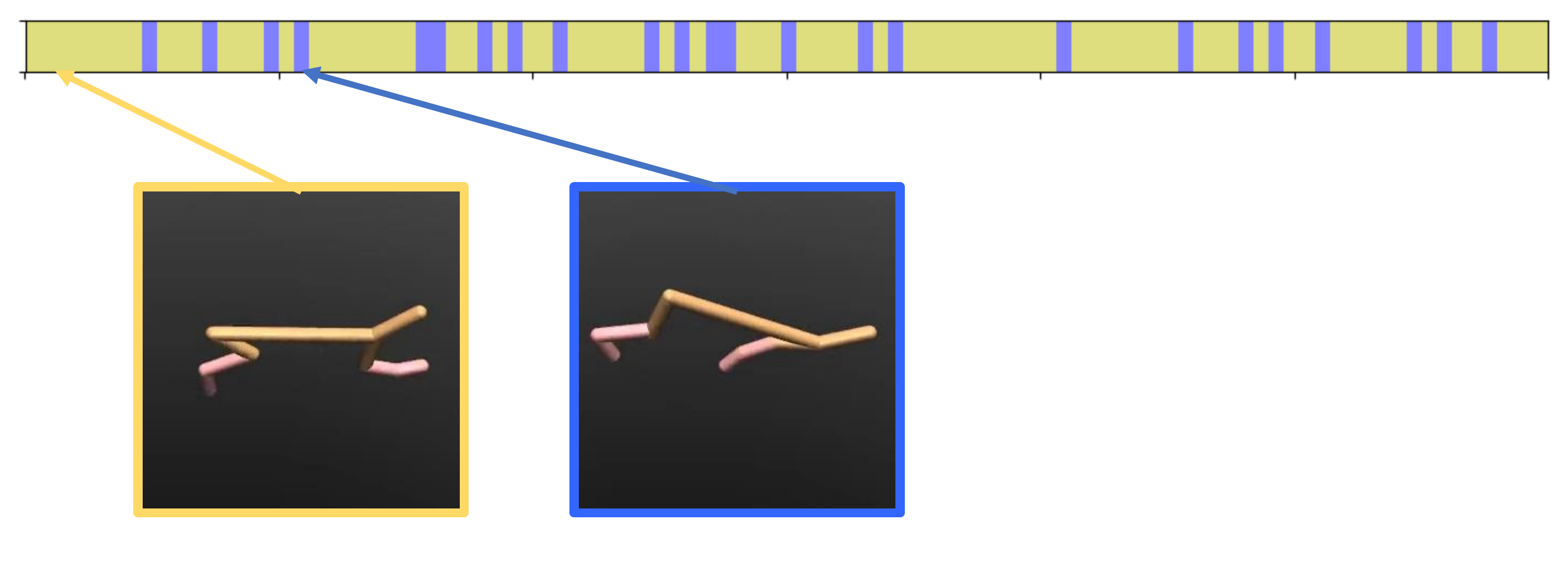}
	\caption{Activation of options over time steps on the HalfCheetah-v1 task using adInfoHRL with two options.}
	\label{fig:option_time_halfCheetah_option2}
\end{figure}

The activation of the options over time is shown in Figure~\ref{fig:option_time_halfCheetah_option2}.
It is clear that one of the option corresponds to the stable running phase and the other corresponds to the phase for recovering from unstable states.

\begin{figure}[]
	\centering
	\begin{subfigure}[t]{0.4\columnwidth}
		\includegraphics[width=0.8\textwidth]{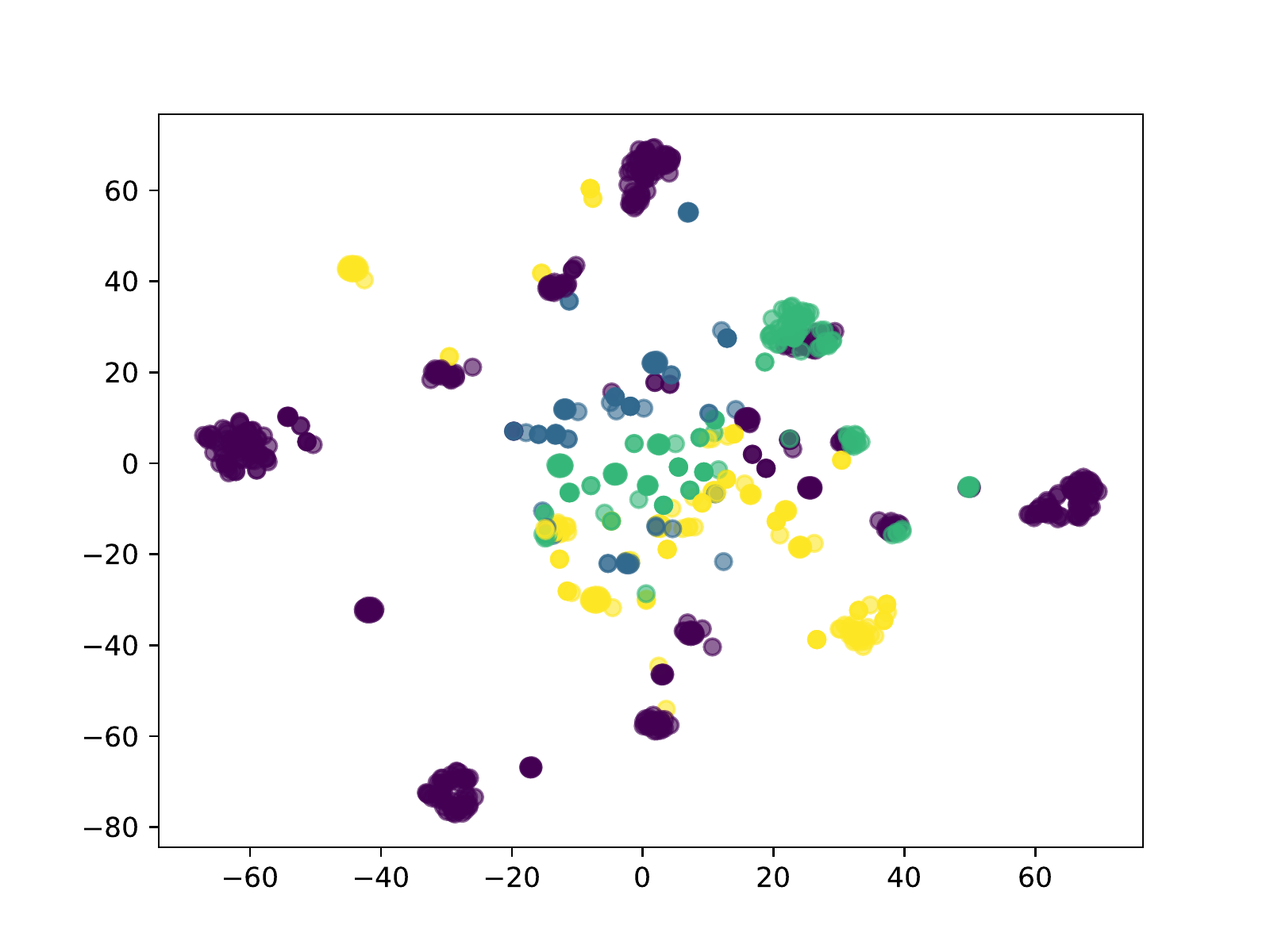}
		\caption{Output of the option network in the state-action space on the Ant-rllab task. }
	\end{subfigure}
	\begin{subfigure}[t]{0.4\columnwidth}
		\includegraphics[width=0.8\textwidth]{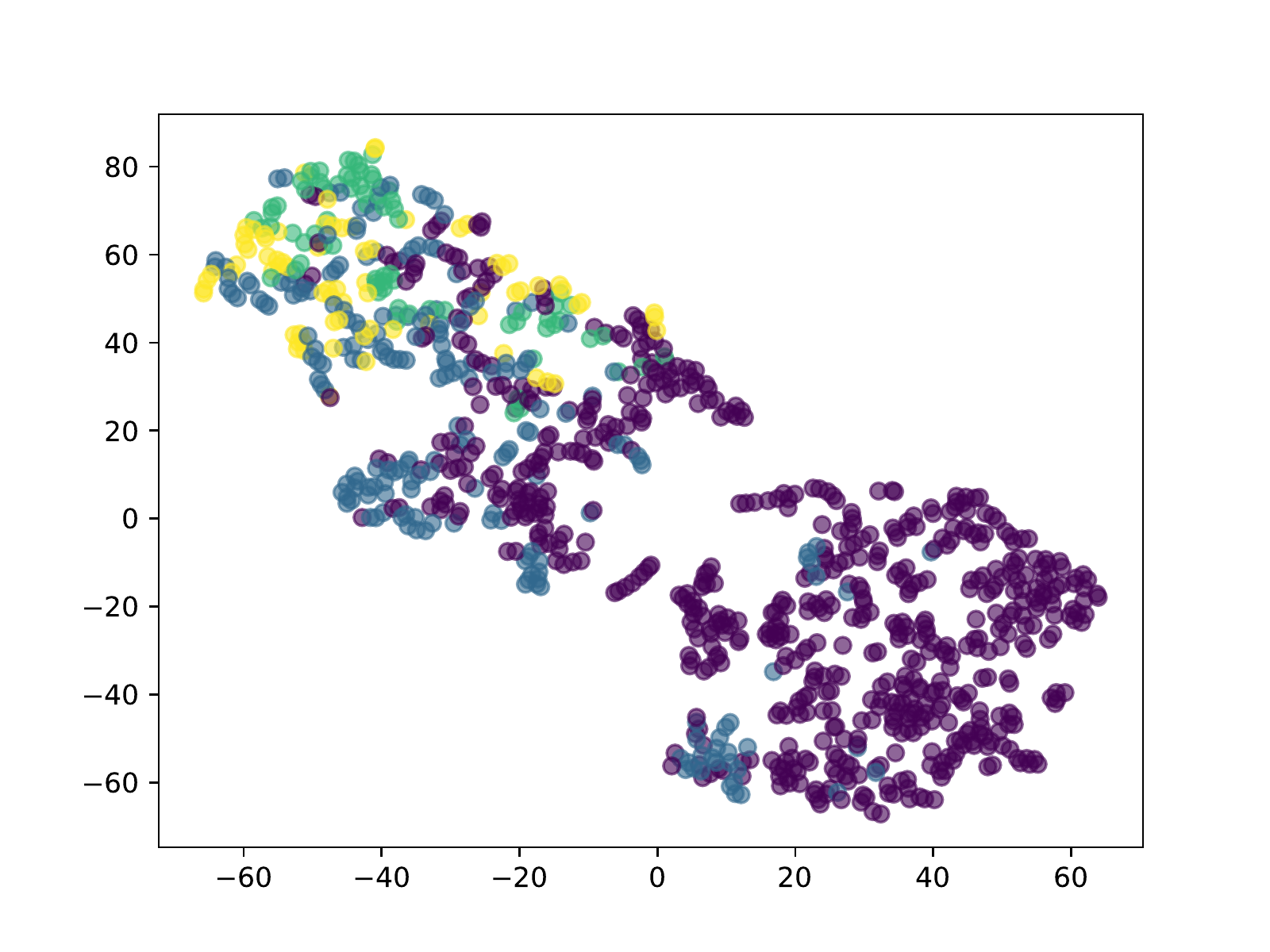}
		\caption{Activation of options in the state space on the Ant-rllab task. }
	\end{subfigure}
	\caption{Distribution of options on Ant-rllab task using adInfoHRL with four options. The dimensionality is reduced by t-SNE for visualization.}
	\label{fig:performance_Ant_rllab}
\end{figure}

\begin{figure}[]
	\centering
	\includegraphics[width=0.9\columnwidth]{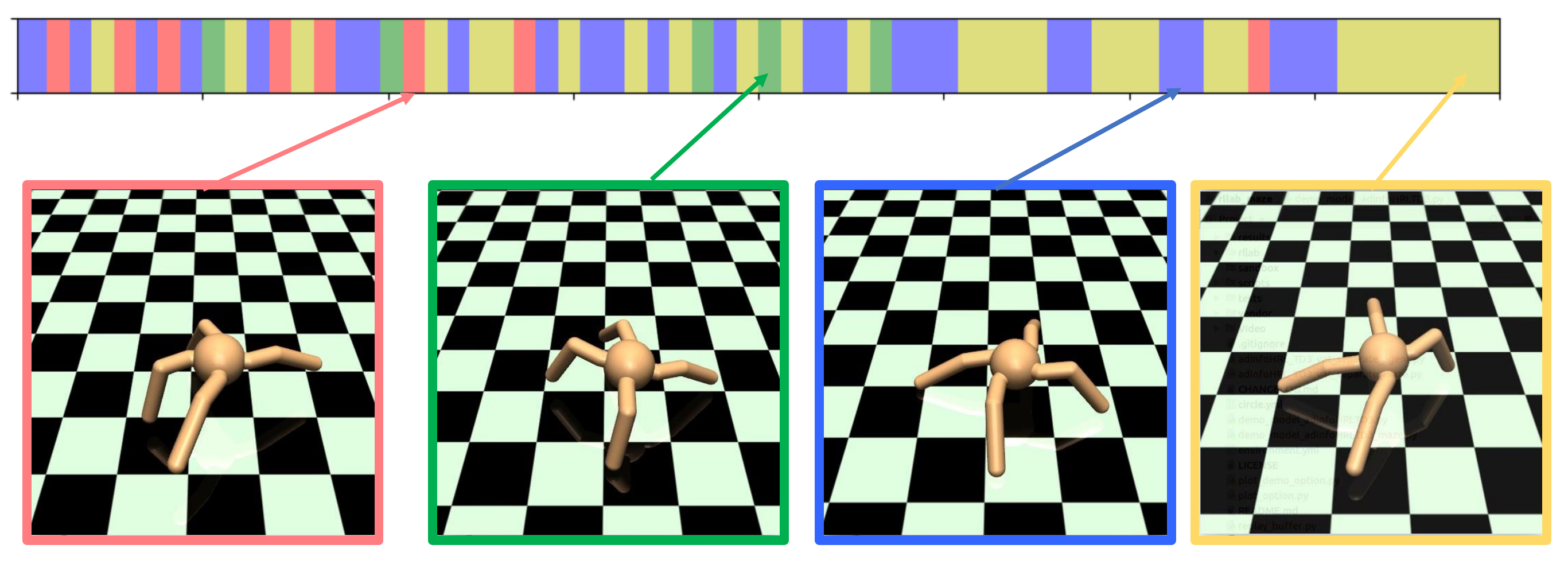}
	\caption{Activation of the options over time steps on Ant-rllab task. Four options are learned.}
	\label{fig:option_time_Ant_rllab}
\end{figure}

The distribution of four options on the Ant-rllab task after one million steps is shown in Figure~\ref{fig:performance_Ant_rllab}.
Four options are activated in the different domains of the state space.
The activation of the options over time  on the Ant-rllab task is shown in Figure~\ref{fig:option_time_Ant_rllab}.
While four options are actively used in the beginning of the episode, two (blue and yellow) options are mainly activated during the stable locomotion.

\begin{figure}[]
	\centering
		\begin{subfigure}[t]{0.33\columnwidth}
		\includegraphics[width=0.8\textwidth]{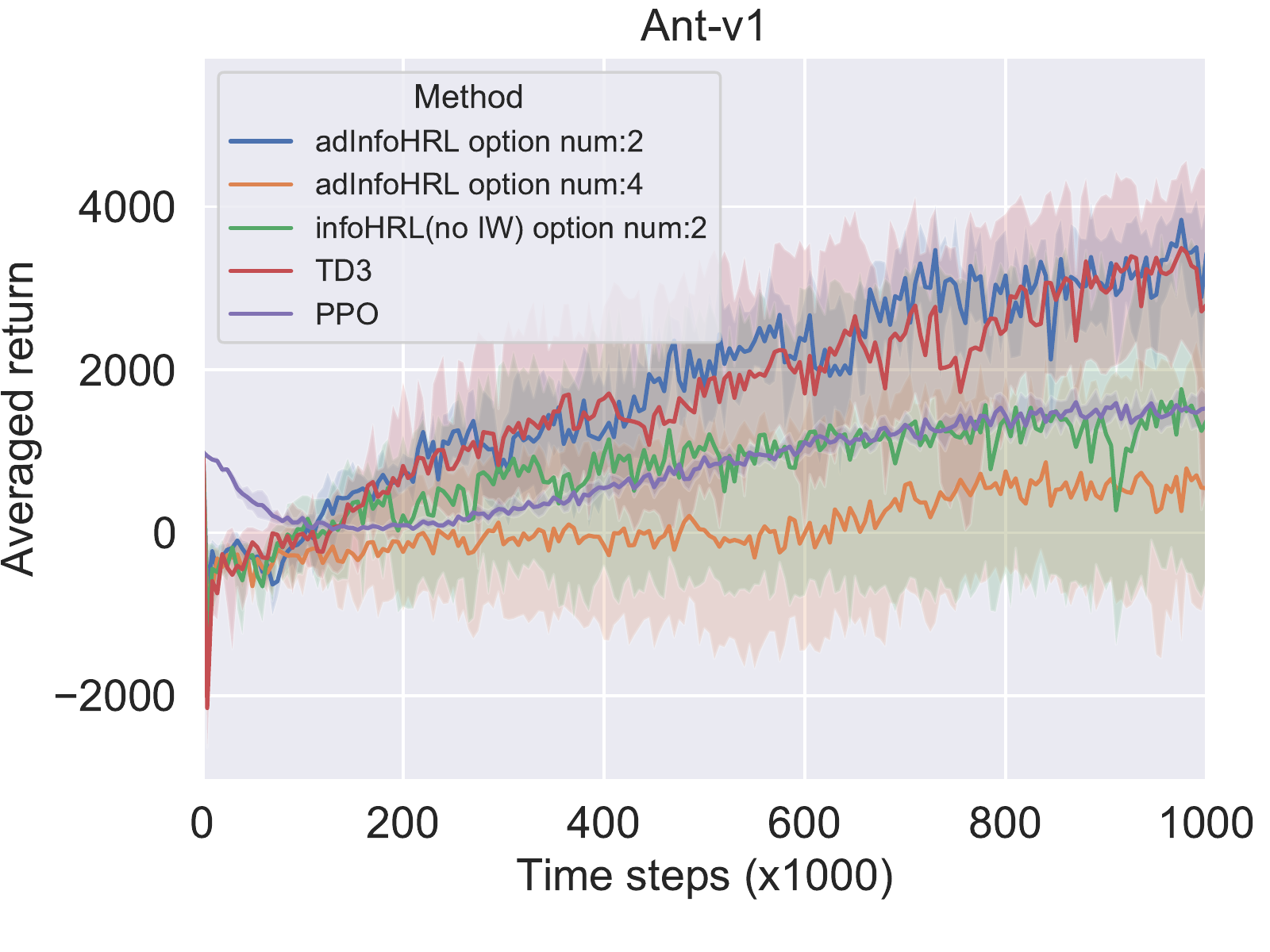}
		\caption{Averaged return on the Ant-v1 task in the OpenAI gym. }
	\end{subfigure}
	\begin{subfigure}[t]{0.32\columnwidth}
		\includegraphics[width=0.8\textwidth]{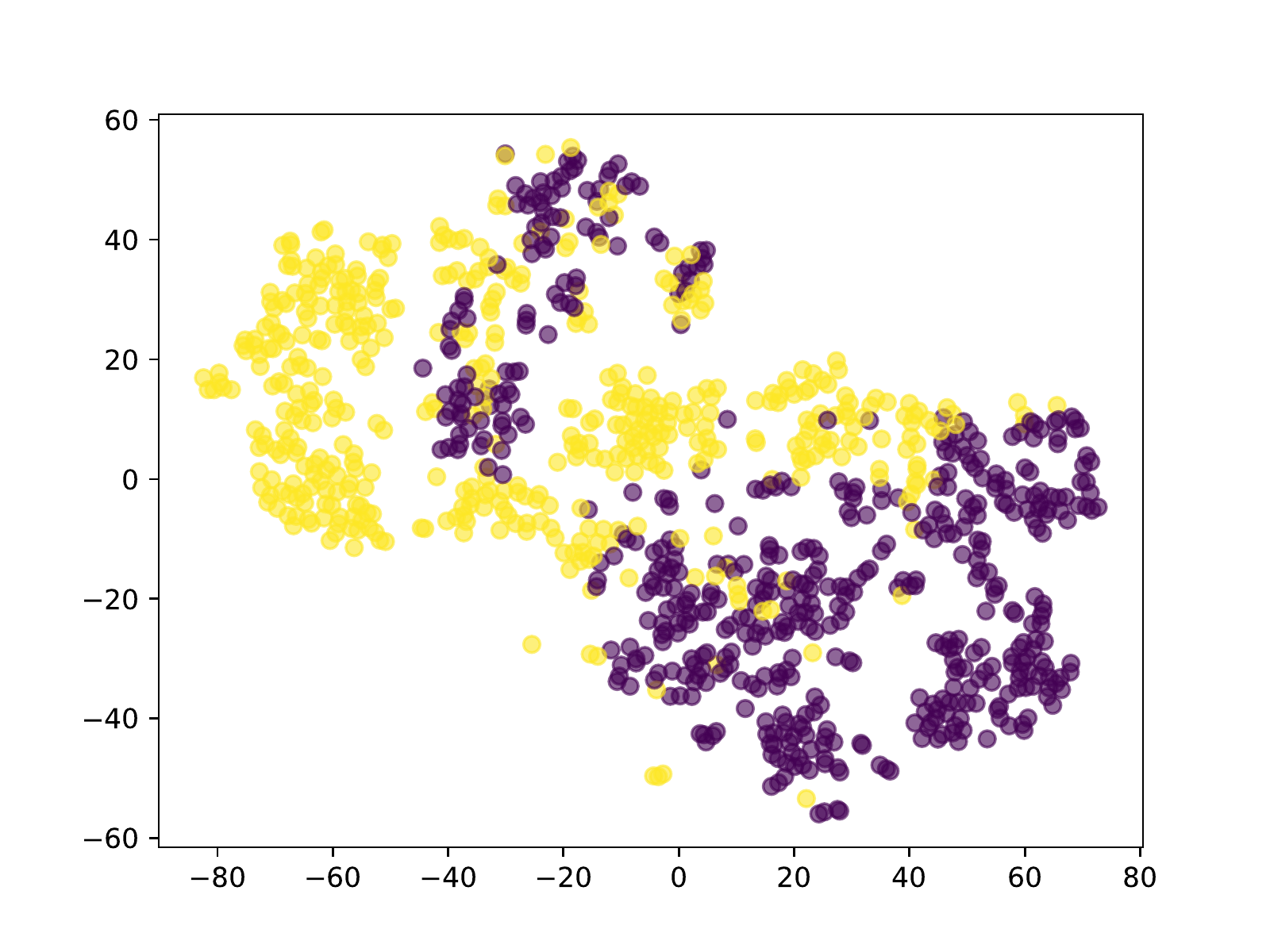}
		\caption{Output of the option network in the state-action space on Ant-v1. }
	\end{subfigure}
	\begin{subfigure}[t]{0.32\columnwidth}
		\includegraphics[width=0.8\textwidth]{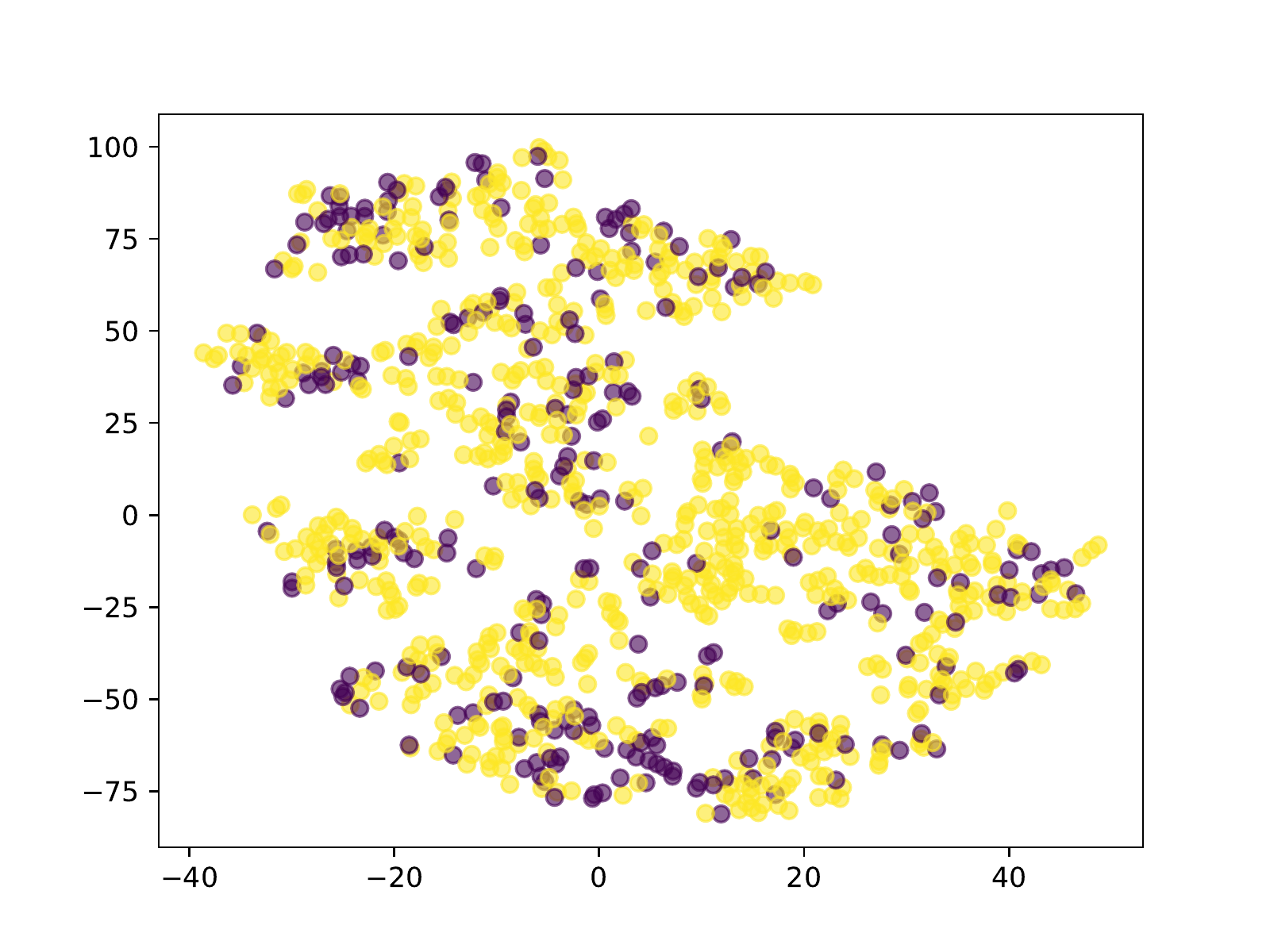}
		\caption{Activation of options in the state space on Ant-v1. }
	\end{subfigure}
	\caption{Distribution of options on the Ant-v1 task using adInfoHRL with two options. The dimensionality is reduced by t-SNE for visualization.}
	\label{fig:performance_Ant}
\end{figure}

\begin{figure}[]
	\centering
	\includegraphics[width=0.9\columnwidth]{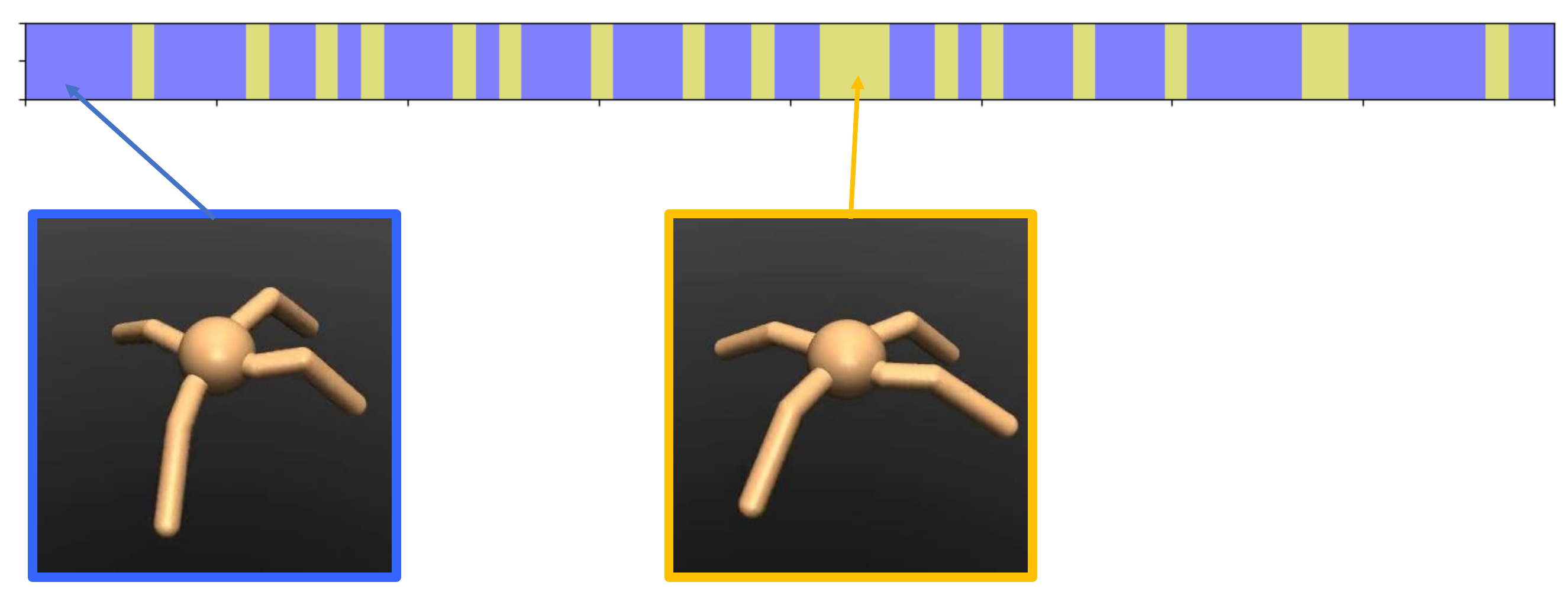}
	\caption{Activation of options over time steps on the Ant-v1 task using adInfoHRL with two options.}
	\label{fig:option_time_ant_option2}
\end{figure}

Since the Ant task implemented in rllab is known to be harder than the Ant-v1 implemented in the OpenAI gym, 
we reported the result of the Ant task in rllab in the main manuscript.
Here, we report the result of the Ant-v1 task implemented in the OpenAI gym.
On the Ant-v1 task, adInfoHRL yielded the best performance with two options.
The performance of adInfoHRL with two options is comparable to that of TD3 on Ant-v1.
This result indicates that the Ant-v1 task does not require a hierarchical policy structure, while a hierarchical policy improves the performance of learning on Ant-rllab.
The distribution of options on Ant-v1 task after one million steps is shown in Figure~\ref{fig:performance_Ant}.
The activation of the options over time is shown in Figure~\ref{fig:option_time_ant_option2}.
It is evident that two option policies on the Ant-v1 task corresponded to different postures of the agent.
%Compared with Walker2d-v1 and HalfCheetah-v1, the activation of two options is not clearly separated while the action-state space is clearly separated.
%The Q-function used for activating options may not be sufficiently trained after 1 million steps due to the high dimensionality of the Ant-v1 task. 

%Recent studies on HRL reported the performance on the same task we used in our experiments.
A recent study on HRL by \citet{Smith18} reported the performance of IOPG on Walker2d-v1, Hopper-v1, and HalfCheetah-v1. 
The study by \citet{Haarnoja18} reported the performance of SAC-LSP on Walker2d-v1, Hopper-v1, HalfCheetah-v1, and Ant-rllab.
A comparison of performance between our method, IOPG, and SAC-LSP is summarized in Table~\ref{tbl:comparison}.
We report the performance after 1 million steps.
It is worth noting that adInfoHRL outperformed IOPG on these tasks in terms of the achieved return,
although we are aware that the qualitative performance is also important in HRL.
AdInfoHRL outperformed SAC-LSP on Walker2d-v1 and Ant-rllab, and SAC-LSP shows its superiority on HalfCheetah-v1 and Hopper-v1.
However, the results of SAC-LSP were obtained by using reward scaling, which was not used in the evaluation of adInfoHRL. 
Therefore, further experiments are necessary for fair comparison under the same condition.

\begin{table}[]
	\caption{Comparison of performance with existing methods after 1 million steps as reported in the literature. 
		For adInfoHRL, we show the mean and the standard deviation of the results from the 10 different seeds. 
		The performance of IOPG and SAC-LSP is from the original papers~\citep{Smith18} and \citep{Haarnoja18}}
	\centering
	\begin{tabular}{lcccc}
		\hline
		%\multicolumn{2}{c}{Part}                   \\
		%\cmidrule{1-2}
		task     & adInfoHRL (two opt.)    & adInfoHRL (four opt.) & IOPG & SAC-LSP \\
		\hline
		Walker2d-v1 & \textbf{3752.1 $\pm$ 442}  &  3404.2$\pm$ 785.6  &  $\approx$ 800 & $\approx$ 3000  \\
		\hline
		HalfCheetah-v1 & 6315.1 $\pm$ 612.8  &  4520.6 $\pm$ 859.3  &  $\approx$ 800 & \textbf{$\approx$ 8000} \\
		\hline
		Hopper-v1 &  1821.7 $\pm$ 626.3 & 1717.5 $\pm$ 281.7   &  $\approx$ 1000 & \textbf{$\approx$ 2500} \\
		\hline
		Ant rllab &  \textbf{1263.2 $\pm$  333.5}  &  683.2 $\pm$ 105.68   &  -- & $\approx$ 500 \\
		\hline
	\end{tabular}
	\label{tbl:comparison}
\end{table}

\end{document}